\documentclass[conference]{IEEEtran}
\usepackage{times}

\usepackage[numbers,square,comma,sort&compress]{natbib}
\usepackage{multicol}
\usepackage[bookmarks=true,pagebackref=true]{hyperref}
\usepackage{multirow}
\usepackage{url}
\usepackage{float}
\usepackage{comment}
\usepackage{mathtools}
\usepackage{amsfonts}  %
\usepackage{cleveref}
\usepackage{graphics}
\usepackage[pdftex]{graphicx}
\usepackage{wrapfig}
\usepackage[font=footnotesize]{caption}
\usepackage{color}
\usepackage{dsfont}
\usepackage[]{mdframed}
\usepackage{algorithm}
\usepackage{algpseudocode}
\usepackage{xparse}
\usepackage{amsmath}
\usepackage{bm}
\usepackage{mathtools}
\usepackage{amssymb}
\usepackage{amsthm}
\usepackage{amssymb}

\usepackage{booktabs}
\usepackage{epigraph}
\usepackage{listings}
\usepackage{xcolor} %
\usepackage{subcaption}
\let\labelindent\relax
\usepackage{enumitem}
\usepackage{tcolorbox}

\usepackage{tikz}

\lstdefinestyle{pythonstyle}{
    language=Python,
    basicstyle=\small\ttfamily,
    keywordstyle=\color{blue},
    stringstyle=\color{red},
    commentstyle=\color{green!50!black},
    morecomment=[l]{\#},
    showstringspaces=false,
    numbers=left,
    numberstyle=\tiny\color{gray},
    frame=single,
    breaklines=true,
    tabsize=4
}

\usepackage{xcolor} %
\usepackage{hyperref} %
\hypersetup{
    colorlinks=true,
}

\usepackage{xargs} %
\usepackage[colorinlistoftodos,prependcaption,textsize=tiny]{todonotes}
\newcommandx{\wrn}[2][1=]{\todo[linecolor=red,backgroundcolor=red!25,bordercolor=red,#1]{#2}}
\newcommandx{\cmt}[2][1=]{\todo[linecolor=blue,backgroundcolor=blue!25,bordercolor=blue,#1]{#2}}

\usepackage{mdframed}

\newcommand{\ba}{\mathbf{a}}

\newcommand{\bo}{\mathbf{o}}
\newcommand{\bq}{\mathbf{q}}
\newcommand{\bg}{\mathbf{g}}
\newcommand{\bI}{\mathbf{I}}

\newcommand{\E}{\mathbb{E}}
\newcommand{\lang}{\ell}

\newcommand{\rawtext}{\hat{\ell}} %
\newcommand{\data}{\mathcal{D}}

\def \Piz {$\pi_0$}

\def \Pizf {$\pi_{0.5}$}
\def \Pizs {$\pi_{0.6}$}
\def \Piszs {$\pi^{*}_{0.6}$}

\def \MethodName {$\pi_{0.7}$}
\def \ModelSymbol {$\pi_{0.7}$}
\IEEEoverridecommandlockouts

\begin{document}

\makeatletter
\let\@oldmaketitle\@maketitle
\renewcommand{\@maketitle}{\@oldmaketitle
    \begin{center}
    \captionsetup{type=figure}
    \includegraphics[width=1.0\textwidth]{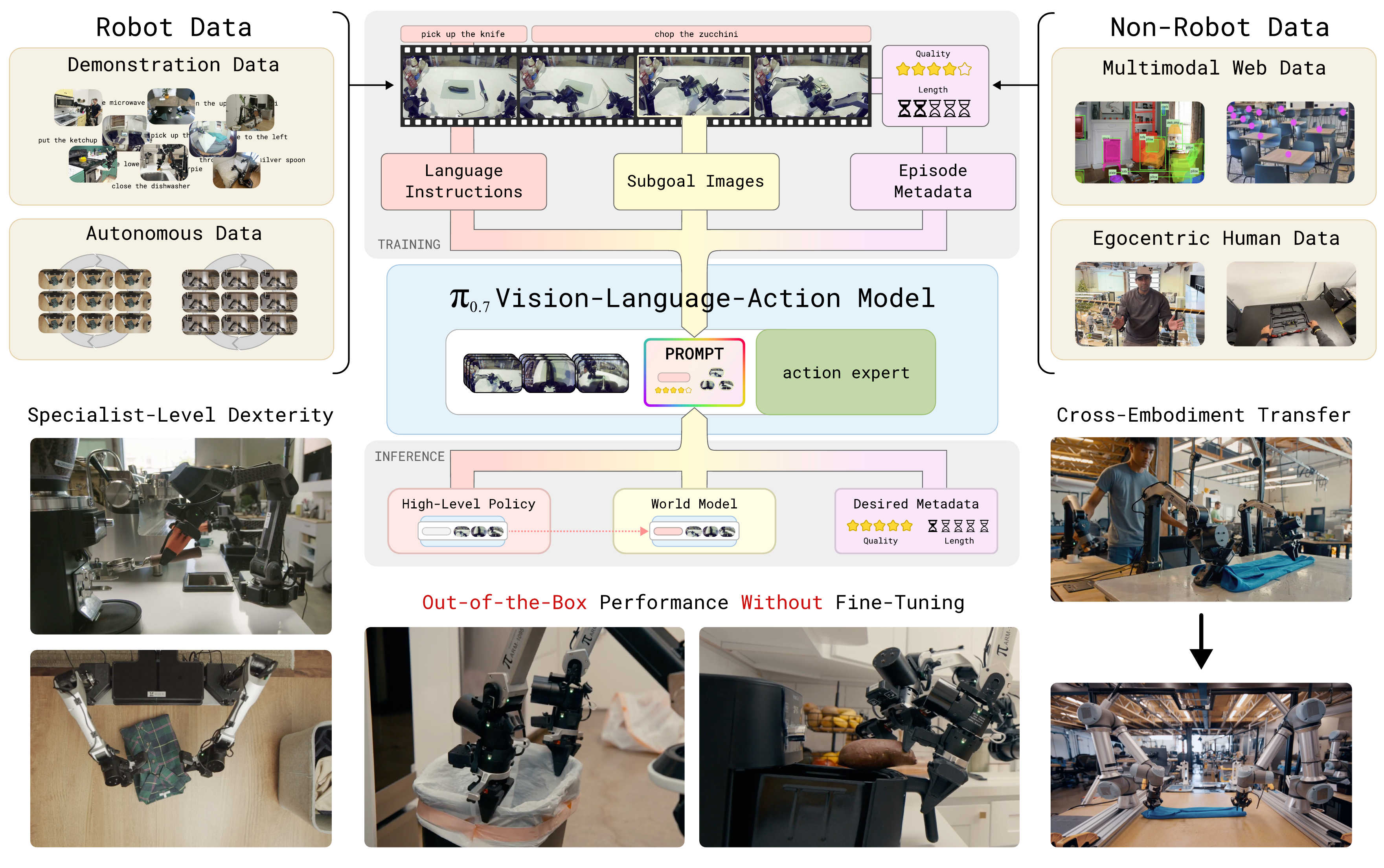}
    \caption{We introduce \MethodName{}, a steerable generalist robot foundation model that can perform dexterous tasks across many tasks, environments, and robots. \MethodName{} is trained with diverse prompts that contain not only the task description, but detailed language, generated subgoal images, and episode metadata. This provides richer context about not only \emph{what} to do, but also \emph{how} to do it, making it possible for \MethodName{} to leverage broad set of both robot and non-robot data and compose the skills in this data in new ways to solve new tasks.}
    \label{fig:teaser}
    \end{center}
}
\makeatother

\title{
\ModelSymbol: a Steerable Generalist Robotic Foundation Model
with Emergent Capabilities
}

\pdfinfo{
   /Author (Physical Intelligence)
   /Title  (@title)
   /Subject (Robot Foundation Models)
   /Keywords (Robot Foundation Models)
}

\author{
\centering
\begin{minipage}{0.95\textwidth}
\centering
\textbf{Physical Intelligence}\\
\vspace{0.2em}
\footnotesize
Bo Ai, Ali Amin, Raichelle Aniceto, Ashwin Balakrishna, Greg Balke, Kevin Black, George Bokinsky, Shihao Cao, Thomas Charbonnier,\\
Vedant Choudhary, Foster Collins, Ken Conley, Grace Connors, James Darpinian, Karan Dhabalia, Maitrayee Dhaka, Jared DiCarlo, Danny Driess,\\
Michael Equi, Adnan Esmail, Yunhao Fang, Chelsea Finn, Catherine Glossop, Thomas Godden, Ivan Goryachev, Lachlan Groom, Haroun Habeeb,\\
Hunter Hancock, Karol Hausman, Gashon Hussein, Victor Hwang, Brian Ichter, Connor Jacobsen, Szymon Jakubczak, Rowan Jen, Tim Jones,\\
Gregg Kammerer, Ben Katz, Liyiming Ke, Mairbek Khadikov, Chandra Kuchi, Marinda Lamb, Devin LeBlanc, Brendon LeCount, Sergey Levine,\\
Xinyu Li, Adrian Li-Bell, Vladislav Lialin, Zhonglin Liang, Wallace Lim, Yao Lu, Enyu Luo, Vishnu Mano, Nandan Marwaha, Aikys Mongush,\\
Liam Murphy, Suraj Nair, Tyler Patterson, Karl Pertsch, Allen Z. Ren, Gavin Schelske, Charvi Sharma, Baifeng Shi, Lucy Xiaoyang Shi,\\
Laura Smith, Jost Tobias Springenberg, Kyle Stachowicz, Will Stoeckle, Jiaming Tang, Jimmy Tanner, Shalom Tekeste, Marcel Torne,\\
Kyle Vedder, Quan Vuong, Anna Walling, Haohuan Wang, Jason Wang, XuDong Wang, Chris Whalen, Samuel Whitmore, Blake Williams,\\
Charles Xu, Sukwon Yoo, Lili Yu, Wuming Zhang, Zhuoyang Zhang, Ury Zhilinsky\\[0.3em]
\normalsize
\url{https://pi.website/pi07}
\end{minipage}
}

\maketitle

\begin{abstract}
We present a new robotic foundation model, called \MethodName{}, that can enable strong out-of-the-box performance in a wide range of scenarios. \MethodName{} can follow diverse language instructions in unseen environments, including multi-stage tasks with various kitchen appliances, provide zero-shot cross-embodiment generalization, for example enabling a robot to fold laundry without seeing the task before, and perform challenging tasks such as operating an espresso machine out of the box at a level of performance that matches much more specialized RL-finetuned models. The main idea behind \MethodName{} is to use diverse context conditioning during training. This conditioning information, contained in the prompt, makes it possible to steer the model precisely to perform many tasks with different strategies. It is conditioned not just on a language command that describes what it should do, but on additional multimodal information that also describes the manner or strategy in which it should do it, including metadata about task performance and subgoal images. This enables \MethodName{} to use very diverse data, including demonstrations, potentially suboptimal (autonomous) data including failures, and data from non-robot sources. Our experiments evaluate \MethodName{} across numerous tasks with multiple robot platforms, on tasks that require speed and dexterity, language following, and compositional task generalization.
\end{abstract}

\IEEEpeerreviewmaketitle

\section{Introduction}

\setlength{\epigraphwidth}{0.42\textwidth}
\epigraph{\textit{I am a part of all that I have met.}}{Alfred, Lord Tennyson, \textit{Ulysses}}

Foundation models work on the principle that generalist capabilities emerge from training on large and diverse datasets. For example, large language models can not only recall facts and semantic knowledge, but they can also compose that knowledge in new ways, solving problems that require unlikely connections, applying user-defined formats (e.g., JSON), and performing chain-of-thought reasoning. 
This kind of \emph{compositional} generalization is arguably the cornerstone of generalist capabilities, but it has proven elusive in the domain of physical intelligence. While robotic foundation models such as vision-language-action models (VLAs) have advanced significantly in size and capability, their ability to generalize to new tasks or recombine skills in new ways has so far been limited. Unlike language models, which can compose different capabilities from their training data to solve new problems, prior VLAs not only lack the ability to solve new tasks, but often struggle to fluently perform all of the instructions they were trained on without task-specific fine-tuning.

In this paper, we present a new model, \MethodName{}, that exhibits strong signs of compositional generalization --- enabling it to follow diverse language instructions, attain performance comparable to more specialized fine-tuned models on dexterous tasks, and even compose these behaviors in new ways. This is enabled by leveraging large and diverse datasets, including data from many robots with diverse strategies, suboptimal data from autonomous execution (including both data from RL post-trained agents as well as failures), and non-robot data from videos of humans performing tasks and general multimodal data from the internet. However, using such data na\"{i}vely does not lead to success: with a diversity of examples that differ in terms of both strategy and task performance, a na\"{i}ve training process would lead to a model that averages together different modes in the dataset and produces suboptimal results. In training \MethodName{}, we address this challenge by annotating the data with detailed context annotations that contain not only information about \emph{what} to do but also \emph{how} to do it and provide this knowledge to the model using a variety of multimodal conditioning signals. In this way, each episode teaches the robot about nuanced concepts and skills that it could use not only to perform the training tasks effectively, but also to compose in new ways to solve new tasks. Our proposed prompt structure includes detailed language labels, strategy metadata, and multimodal information such as subgoal images. This allows us to resolve the ambiguity in large and diverse datasets, learn from suboptimal behaviors without hurting performance, and obtain broad generalization across instructions, embodiments, and environments.

The idea that detailed prompts or context can improve the performance of foundation models has been explored in other fields. For example, models for image and video generation utilize \emph{prompt expansion} to produce high-quality generations. Our approach has many parallels to such methods. However, in robotics, simply \emph{captioning} the data with more detailed text is not enough --- the details that determine task success and proficiency might be more subtle (e.g., information about the overall quality of the episode), or might simply be hard to express with language alone (e.g., the particular appearance of a cleanly folded t-shirt). Therefore, in addition to using more detailed text, our model adds a range of additional metadata to the prompt, as shown in Fig.~\ref{fig:teaser}, including information about episode quality (strategy metadata), the control modality used by the robot, and subgoal images. Some of this information can be provided or omitted at test time, but including it in training results in a model that can more effectively compose the concepts that it was trained on and exhibit a variety of emergent capabilities.

In our evaluation, we show that \MethodName{} exhibits a number of capabilities that go beyond prior robotic foundation models:
\begin{itemize}[leftmargin=10pt]
    \item \textbf{Out-of-the-box performance:} \MethodName{} can reliably perform highly dexterous, long-horizon tasks such as using an espresso machine, folding laundry, taking out a trash bag, folding a box, and peeling vegetables, without any task-specific post-training and in a variety of environments.
    \item \textbf{Instruction generalization:} \MethodName{} can follow a diverse set of language instructions in unseen environments and demonstrates robust generalization to complex, unseen language references. For example, \MethodName{} can follow a diverse set of open-ended instructions in entirely unseen kitchen and bedroom environments.
    \item \textbf{Cross-embodiment generalization:} \MethodName{} can enable zero-shot cross-embodiment transfer, making it possible to transfer dexterous tasks such as folding a t-shirt to a robot that was never trained to perform any laundry folding task, matching the performance of expert operators teleoperating the robot on their initial attempts.
    \item \textbf{Compositional task generalization:} \MethodName{} can be instructed to perform new tasks by composing skills in previously unseen ways. For example, we can prompt \MethodName{} to use new kitchen appliances, such as loading a sweet potato into an air fryer, or prompt it to perform tasks in new ways.
\end{itemize}
\noindent Through ablation and scaling studies, we also empirically demonstrate that there is a strong synergy between diverse datasets and detailed contexts: our approach enables learning from mixed-quality data and non-standard data sources without hurting the model performance, and diverse data boosts the model performance when detailed context information is provided during training. 

\section{Related Work}
\label{sec:related_work}

\noindent \textbf{Generalist robot manipulation policies.} There is a large body of work studying the development of generalist robot policies.
These generalist policies are sometimes trained from scratch~\cite{brohan2022rt,reed2022gato,team2024octo,liu2024rdt1b,wang2024hpt,lbmtri2025}, but are more commonly initialized using pre-trained vision-language models~\cite{rt22023arxiv,open_x_embodiment_rt_x_2023,kim2024openvla,black2024pi_0,wen2024tinyvlafastdataefficientvisionlanguageaction,zhen20243dvla,geminirobotics2025,black2025pi05,zheng2025x,jiang2025galaxea,li2024roboflamingo,li2024cogact,qu2025spatialvla,bjorck2025groot,zawalski2024ecot,agibotworld2025, zhou2025chatvla} or pre-trained video generation models~\cite{kim2025cosmospolicy,mimicvideo,ye2026dreamzero,wu2024gr1,cheang2024gr2}.
Various works have developed architectural components of VLAs such as memory~\cite{zheng2024tracevla,sridhar2025memer,shi2025memoryvla,lin2025onetwovla,fang2025sam2act,li2025cronusvla,zhang2025ta,jang2025contextvla,torne2026mem}, hierarchy for long-horizon planning~\cite{ahn2022saycan,liang2023codepolicies,shi2025hi,black2025pi05,geminirobotics2025}, and goal image conditioning~\cite{zhao2025cot}. We develop a VLA model that incorporates all three of these components in a single model, building on top of the $\pi_{0.6}\texttt{-MEM}$ architecture~\cite{pi06model,torne2026mem}.
While generalist policies are most often trained on robot demonstration data, prior works have shown how to derive benefits from incorporating web data~\cite{rt22023arxiv}, egocentric videos of humans~\cite{ye2024latent,lin2025physbrainhumanegocentricdata,kareer2025emergence,li2025latbotdistillinguniversallatent,yang2025egovlalearningvisionlanguageactionmodels,mimicvideo,luo2026beingh05scalinghumancentricrobot,zhang2026clapcontrastivelatentaction}, and autonomous robot experience~\cite{pistar06,xu2024rldg,xiao2025self} into pretraining. We incorporate all of these data sources and find that the combination of diverse data with detailed prompting yields a model with strong signs of compositional generalization and performant out-of-the-box behavior.

\noindent \textbf{Generalization across tasks and embodiments.} Much prior work has aimed to learn robot policies that generalize not only to different environments, objects, and backgrounds, but also to entirely new tasks and embodiments.
Often, this is done by leveraging human video data, either for general representation learning~\citep{nair2023r3m,ma2022vip,xiao2022masked,bhateja2023robotic,zhou2021manipulator,bharadhwaj2023visual}, by directly supervising with human motions~\citep{chen2026dexterous,shaw2023videodex,bharadhwaj2023zero,bahl2022human,bahl2023affordances,kareer2025egomimic,kareer2025emergence,shi2025learning}, or by extracting 2D point tracks~\citep{bharadhwaj2024track2act,vecerik2024robotap,wen2023any,gu2024rttrajectory}.
Other work has aimed to improve generalization by directly leveraging Internet pre-trained foundation models during training or inference~\citep{kapelyukh2023dall,mandi2022cacti,chen2023genaug,yu2023scaling,stone2023open,driess2023palme,jiang2023vima}.
With the increasing availability of large, cross-embodiment robot datasets~\citep{collaboration2023open}, there has also been work on explicitly improving cross-embodiment transfer between robots~\citep{yang2026data,Doshi24-crossformer,yang2024pushing,zha2026lap,grover2025enhancing,ai2025towards,he2025scaling}.
Rather than leveraging existing datasets, some works have proposed specialized hand-held devices that can be used to collect data that can then generalize to various robot embodiments~\citep{chi2024universal,young2021visual}.
In this work, we find that the right prompting allows our model to leverage diverse robot, human, and Internet data to achieve strong generalization across tasks and embodiments.

\noindent \textbf{Prompting robots with subgoal images.} A core architectural component of our model relative to $\pi_{0.6}\texttt{-MEM}$ is to allow the model to be prompted using goal images, including generated subgoal images. Conditioning robot manipulation policies on goal images and videos has been explored in a large body of work. Some of these works utilize user-provided images~\cite{pathak2018zero,chebotar2021actionable,bousmalis2023robocat,myers2023grif}, while others condition the policy on generated goal images from a separate model~\cite{nair2018visual,nair2020hierarchical,black2023zero,ko2024learning,kim2025uniskill,liang2025dreamitate,zhang2026foreact,du2024vlp} or in a chain-of-thought fashion~\cite{zhao2025cot}. Alternatively, image and video generation can be integrated into policy training objectives~\cite{liang2025video,kim2025cosmospolicy,mimicvideo,yuan2026fast,ye2026dreamzero,du2023unipi} to improve policy representations and produce more generalizable actions.
We view our contribution as complementary to these works: we do not aim to propose a new architecture or model design, so much as a methodology for enabling VLAs to utilize more diverse data sources, together with an empirical analysis showing that it leads to strong indications of compositional generalization.
To our knowledge, our empirical results go significantly beyond the quantitative improvements reported in prior works, showing zero-shot transfer of dexterous skills like laundry folding to a different robot and generalization to novel object interactions such as operating an air fryer.

\section{Flow-Based Vision-Language-Action Models}
\label{sec:prelim}

VLAs are trained by starting from a pre-trained vision-language model (VLM) backbone, and adapting it for robotic control. The training dataset $\data$ contains robot trajectories, which are sequences of observations $\bo_t$ and actions $\ba_t$. The observations $\bo_t = [\bI_t^1, \ldots, \bI_t^n, \bq_t]$ consist of $n$ camera images $\bI_t^i$ and the joint configuration of the robot $\bq_t$, while the actions $\ba_t$ consist of joint or end-effector commands. 

VLAs are typically trained to predict an \emph{action chunk}, corresponding to a short trajectory of future actions $\ba_{t:t+H}$, based on a recent history of observations $\bo_{t-T:t}$ (often a shorter horizon of actions, $\hat{H} < H$, is executed).
The action chunk can be generated by an ``action expert'', a smaller transformer that attends to the VLM backbone and enables fast inference at runtime. The action expert typically uses a flow matching~\citep{lipman2022flow} (or diffusion) objective that captures the multi-modality of the robot actions. To learn effective representations, our model also uses the knowledge insulation (KI) training recipe~\cite{driess2025knowledge}: the VLM backbone is supervised with FAST tokens \cite{pertsch2025fast}, and while the action expert attends to all of the activations in the VLM backbone, gradients from the action expert do not flow into the VLM backbone, such that the VLM is trained via the comparatively stable discrete cross-entropy loss.

In addition to the observation and action, each training example for the VLA is accompanied by 
a \emph{prompt} or 
\emph{context}, which we denote with $\mathcal{C}_t$. Conventionally this corresponds to a language instruction $\lang_t$, such that $\mathcal{C}_t = (\lang_t)$, provided by a human annotator (e.g., ``clean up the kitchen'').

In designing \MethodName{}, we explore how additional information added to the context for each training example can enable learning from diverse and heterogeneous datasets (including suboptimal behaviors and failures). As we will show, training with this data leads to a model with greater robustness and dexterity, and makes it possible for the model to generalize more broadly. The training objective for the VLA $\pi_\theta$ corresponds to an approximate log-likelihood given by
\begin{equation}
\max_{\theta} \; \E_{\data}
\left[\log \pi_{\theta}(\ba_{t:t+H} \mid \bo_{t-T:t}, \mathcal{C}_t)\right].
\label{eq:vla_il}
\end{equation}
Note that a flow matching action expert optimizes an approximate lower bound rather than a closed form log-likelihood \cite{black2024pi_0}. The dataset $\data$ typically consists of high-quality human demonstration trajectories; however, as mentioned, we use a broader dataset, which includes failed episodes and suboptimal autonomous rollouts, as well as other data sources such as egocentric human video data. We will show how using a sufficiently detailed and informative context $\mathcal{C}_t$ makes it possible to incorporate such diverse data and, perhaps surprisingly, even results in better policy performance and generalization.

\section{\texorpdfstring{\MethodName{}}{pi0.7} Overview}
\setcounter{figure}{1}
\begin{figure*}[t]
    \centering
    \includegraphics[width=\textwidth]{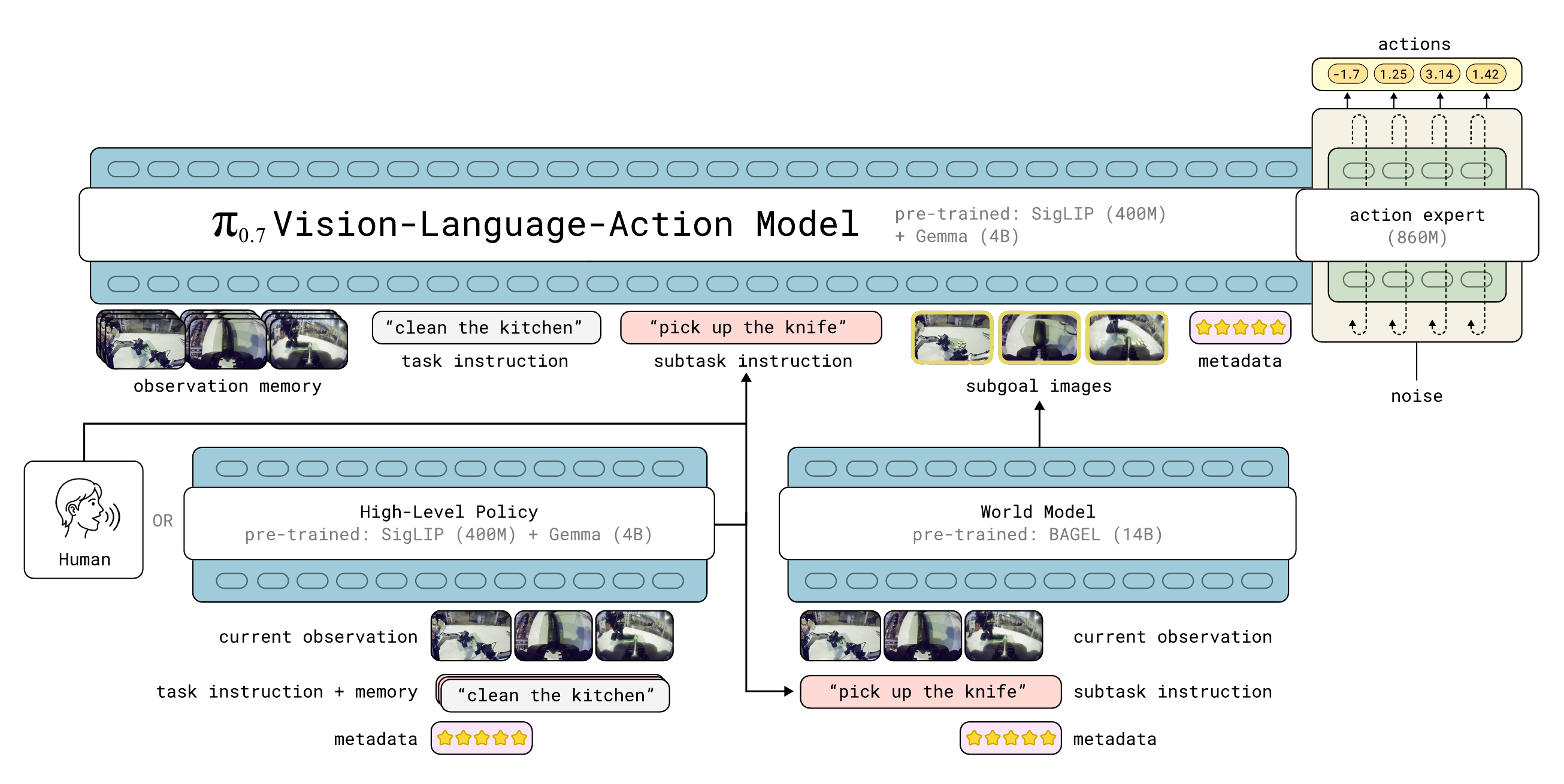}
    \caption{\textbf{Architecture overview.} The \MethodName{} model is a 5B-parameter VLA consisting of a 4B VLM backbone, a \texttt{MEM}-style video history encoder and a 860M parameter action expert. The model's context includes multiple distinct modalities, including language commands, episode metadata that describes the data quality and strategy, and multimodal inputs such as subgoal images. At runtime, the language commands are produced by a high-level semantic policy based on the same architecture, and the subgoal images are produced by a lightweight world model based on the BAGEL image generation model~\cite{deng2025emerging}.}
    \label{fig:arch}
\vspace{-10pt}
\end{figure*}

\MethodName{} is our newest robotic foundation model that builds on the existing VLA architecture from \Pizs{}~\cite{pi06model} and the \texttt{MEM} memory system~\cite{torne2026mem} and extends it with multi-modal context conditioning. The model consists of a VLM backbone initialized from the Gemma3 4B-parameter VLM \cite{gemmateam2025gemma3technicalreport} (including a 400M-parameter vision encoder),
and a flow matching action expert with 860M parameters. The model has about 5B total parameters. The vision encoder is also initialized from Gemma3 and follows the design of the \texttt{MEM} video history encoder~\cite{torne2026mem}, applying both temporal and spatial compression over history observations and outputting a fixed number of tokens for any number of history frames. An overview of the model architecture is provided in Fig.~\ref{fig:arch}, and Sec.~\ref{subsec:architecture} describes the architecture in more detail. 

Our previous models, \Piz{}, \Pizf{}, and \Pizs{}, use a short textual description of the task as the context. In training \MethodName{}, we expand the context to include additional information and modalities: more expressive language commands, episode metadata, and subgoal images, making it possible to train on diverse and potentially suboptimal data.

\section{Diversifying the Prompt}
\label{sec:conditioning}

In this section, we describe each part of the prompt contained in the context $\mathcal{C}_t$ used by \MethodName{}. The model is trained to handle prompts that contain each of these components, though it is trained with each component randomly dropped out so that it can also handle any subset, providing flexibility at test time.

\subsection{Subtask instructions}
\label{subsec:hl}
Following \Pizf\ \cite{black2025pi05}, we include intermediate, higher-level text that captures the \textbf{next semantic subtask} as part of the prompt in addition to the overall textual task description $\ell_t$ (e.g., ``clean the kitchen''). We denote this intermediate text by $\rawtext_t$ (e.g., ``open the fridge door''). During inference, $\rawtext_t$ may be produced by a learned high-level policy or a human (or be omitted) and may change over time.
We collect data from a diverse set of tasks and scenarios, and then annotate the segments with detailed textual descriptions.

Conditioning the model on the semantic subtask also enables us to \textbf{verbally coach} the model step-by-step. Since the model is trained to follow diverse language instructions, it can follow the live instructions from the human in a new task, e.g., loading a sweet potato into an air fryer (Fig.~\ref{fig:air_fryer_coaching}). After coaching we can take the verbal coaching data to finetune \MethodName{} as a high-level policy that maps the robot observations, task specification, and history of past subtask instructions to the new subtask instruction (Fig.~\ref{fig:arch} bottom left). This high-level policy then guides the robot to perform the task fully autonomously.

\subsection{Subgoal images}

While subtask instructions are effective at conveying the high-level intent of the task, they may lack details that matter for execution --- e.g., ``open the fridge door'' does not specify how the robot arm should grasp the handle. Subgoal images address this by depicting the desired near-future state of the scene in images, providing a richer specification of \emph{what the world should look like} after successful progress of the task.

We consider \textbf{multi-view subgoals} $\bg_t = [G_t^1,\ldots,G_t^n]$, where $G_t^i$ is the desired near-future image for camera $i$. Multi-view subgoals simultaneously specify environment- and object-centric outcomes (often easiest in the base view) and arm/gripper outcomes (often easiest in wrist views), improving spatial grounding for control.

At runtime, the subgoal images are produced by a \textbf{lightweight world model}, which takes in the same subtask instruction $\rawtext_t$ as the main model, but benefits from web-scale pre-training on videos and image editing tasks and is thus capable of generalizing to diverse tasks and scenarios. Generated subgoal images that are grounded in the robot's current observation can often more clearly disambiguate the objective for the policy than a language instruction, resulting in improvements in language following and generalization.
We denote this model as $g_\psi$ and it is trained with the objective
\begin{equation*}
\max_{\psi}\;
\E_{\data_g} %
\left[
\mathcal{L}_\text{CFM}
\left(
\bg_t^{\star},\; g_\psi(\bo_t,\rawtext_t,m)
\right)
\right],
\end{equation*}
where $\mathcal{L}_\text{CFM}$ is the standard flow matching loss~\citep{lipman2022flow}, $\bg^{\star}_t$ is the future subgoal image, $m$ is the episode metadata from Sec.~\ref{subsec:metadata}, and the dataset $\mathcal{D}_g$ is a subset of segments from Sec.~\ref{subsec:hl} that are annotated with especially high-quality subtask labels $\rawtext_t$.
The image frames at the end of the segments serve as the ground-truth subgoal, i.e., $\bg_t^{\star} = \bo_{t_{\mathrm{end}}}$. 

Following SuSIE~\citep{black2023zero}, our world model is initialized using an off-the-shelf image generation and editing model with web-scale pre-training. We initialize from BAGEL~\cite{deng2025emerging}, a 14B mixture-of-transformers model capable of image understanding, editing, and generation.
By augmenting our world model training with web data, non-robot data sources such as egocentric human videos, and other video data, we can acquire semantic and physical concepts from these other data sources and then transfer them into \MethodName{} via subgoal images. Implementation details are in Appendix~\ref{app:gg_impl}.

\begin{figure*}[t]
    \centering
    \includegraphics[width=0.75\linewidth]{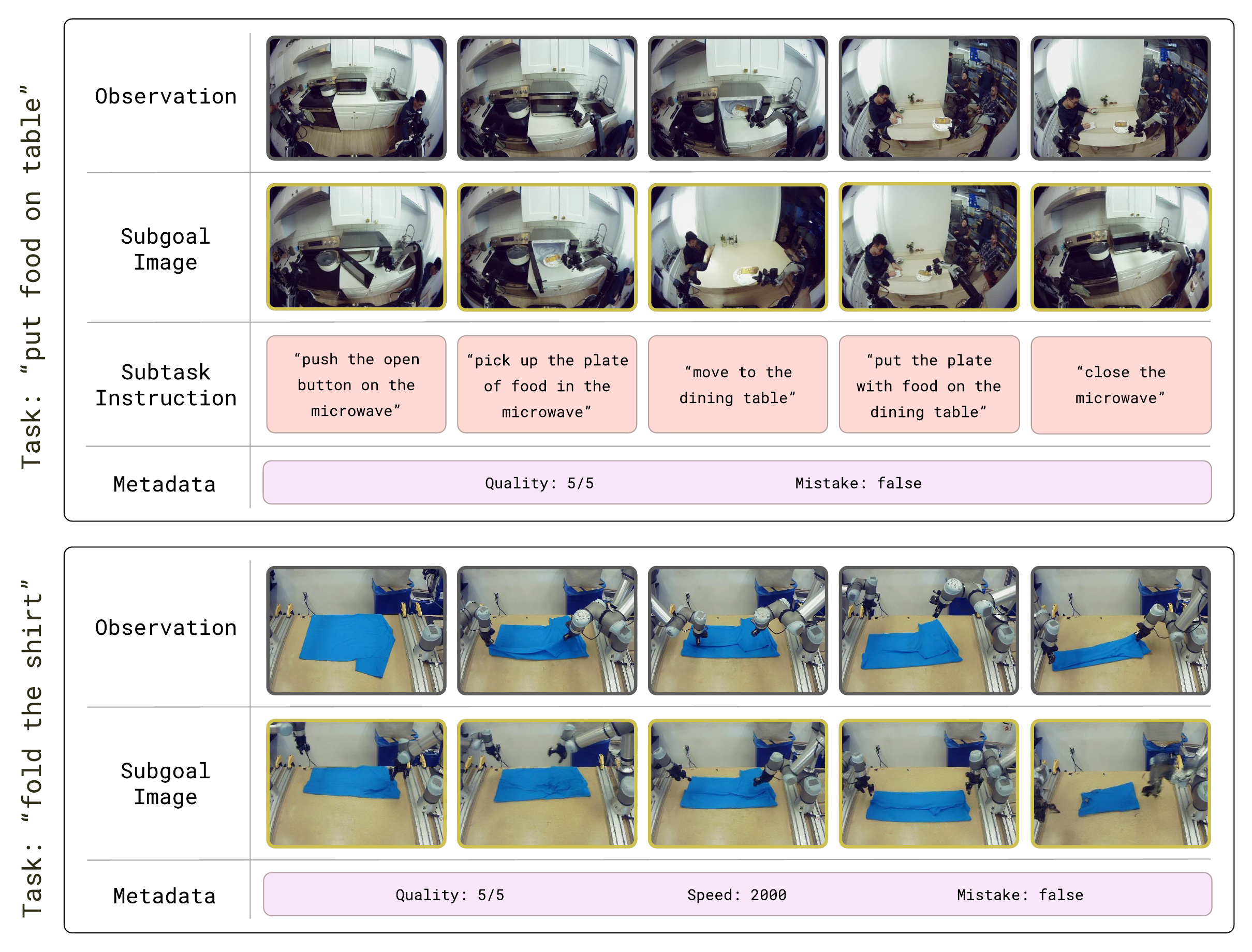}
    \caption{\textbf{Prompt overview.} \MethodName{} uses diverse modalities of context in the prompt, including: subtask instructions, subgoal images, and episode metadata. We train the model with dropout for each component, and then prompt the model flexibly combining modalities. For example, when using the UR5e bimanual manipulator to fold a shirt, we use subgoal image and metadata prompting.}
    \label{fig:prompt}
\vspace{-8pt}
\end{figure*}

\subsection{Episode metadata}
\label{subsec:metadata}

A key goal in expanding the context provided to the model is to train on a broader, more diverse dataset of trajectories. Instead of just using high-quality demonstration data, \MethodName{} leverages lower quality demonstrations (including failures) and even autonomous data from prior models. Since we still want \MethodName{} to perform the task as well as possible at test time, we need to appropriately label these diverse trajectories with information about \emph{how} the task was performed so that the model can correctly contextualize them. To this end, we add a variety of ``episode metadata'' information to the context with attributes of the given training episode. We denote the set of metadata $m$, which may contain various labels including

\begin{itemize}[leftmargin=10pt]
    \item \textbf{Overall speed}: the length of the episode in timesteps. We discretize the values in an interval of 500 steps, i.e., values between 1750 and 2250 are binned to ``2000 steps''. Often faster speed also corresponds to higher quality, e.g., the episode has fewer mistakes.
    \item \textbf{Overall quality}: task execution quality expressed as a score between 1 and 5, with 5 being the highest quality.
    \item \textbf{Mistake}: label indicating whether the robot made a mistake within a given action segment (e.g., failing to grasp an object or performing the wrong subtask). These labels are provided by humans coarsely annotating our data.
\end{itemize}

The \MethodName{} model is thus trained with ground-truth episode speed and manual annotations of the episode quality and mistake segments from a diverse data mixture. The data diversity (e.g., episodes of varying speed) provides the necessary signals for the model to learn to correlate such metadata with the target action.
At runtime, the model can then be instructed to perform the task at high speed, with high quality, and without mistakes, through metadata prompting.

\subsection{Control mode}

We also consider using different control modes for the low-level action execution. Specifically, we include both \textbf{joint-level and end-effector} actions during training and use a text identifier $c \in \{\texttt{joint}, \texttt{ee}\}$ to designate the control mode in the prompt. Then at runtime, we can pick the control mode depending on the task. 

\subsection{Full prompt and training details}

\begin{tcolorbox}
\footnotesize
\texttt{<Multi-view observation><Multi-view subgoals> Task: peel vegetables. Subtask: pick up the peeler. Speed: 8000. Quality: 5. Mistake: false. Control Mode: joint.<Proprioception> }
\end{tcolorbox}
Combining all of the context information together, the example above illustrates a potential prompt that may be provided to the model.

During training, we randomly drop out each part of the prompt, which provides \MethodName{} with the flexibility to use any subset of the prompt components at test time (e.g., running with or without subgoal images).
First, we find that the model trains significantly faster when given the subgoal images --- the action prediction task essentially becomes an ``inverse dynamics'' problem inferring the robot action between the current and future frames. Thus we only add the visual subgoal images to 25\% of the examples in each batch in training. Among the examples with subgoal images we also drop out the subtask instruction $\rawtext_t$ 30\% of the time as often visual subgoal can substitute the equivalent textual subtask description in richer details. For episode metadata, we drop it entirely 15\% of the time, and additionally each component (overall speed, overall quality, and mistake label) is dropped with 5\% probability individually. We do not apply dropout for the control mode.

\section{The \texorpdfstring{\ModelSymbol{}}{pi0.7} Model and Training Recipe}
\label{sec:method}

We now discuss how we incorporate the different context in \ModelSymbol{} model by training on diverse data, as well as the details about the model architecture, training, and inference.

\subsection{Training datasets}

The training dataset for \MethodName{} consists of demonstration data for a wide range of tasks with many different robot platforms (both static and mobile, with single arm or bimanual) in diverse environments (in-house lab-like and home-like environments, and in-the-wild home environments), autonomous data from a large amount of policy evaluations, human interventions within policy rollouts, open-source robot datasets, egocentric human video data, and auxiliary non-robot data sources from the web, including object localization and attribute prediction, visual question answering, and text-only prediction. We also include video-language tasks including video captioning of in-house robot data and from the web.

In a significant departure from classic VLA training pipelines, we make heavy use of suboptimal robot data in training. This includes both lower quality demonstrations (failure episodes or success episodes with a substantial amount of mistakes) and data collected by prior versions of our models during model evaluation experiments\footnote{We exclude autonomous data collected in any generalization-focused evaluation task (including ones in Sec.~\ref{sec:experiments}) from training.}. For example, we use data collected by the \Piszs\ model during RL training as additional examples, effectively allowing \MethodName{} to distill their behavior. Incorporating the episode metadata into the context allows our model to effectively use all of this evaluation data and, as we will see in Sec.~\ref{sec:dexterity}, enables it to attain similar performance as models that are specialized for high performance on individual tasks with RL. This corresponds to a kind of ``distillation'' process, where the generalist \MethodName{} model can inherit the capabilities of RL-trained specialists. Suboptimal data also diversifies the possible states and scenarios in a given task and leads to stronger robustness, enabling the model to even sometimes outperform RL-trained or generally, single-task post-trained policies, in highly dexterous tasks.

\subsection{Model architecture}
\label{subsec:architecture}

The major architectural modifications of \MethodName{} model compared to previous \Pizf{} and \Pizs{} models include the use of the history vision encoder from \texttt{MEM} \cite{torne2026mem} and visual subgoal images in the context.
The model takes as input up to four camera images (front view, two wrist views, and optionally rear view), each with up to six history frames, and up to three subgoal images (omitting the rear view). The history frames are processed through the vision encoder and compressed to the same number of tokens as a single frame; subgoal images are processed through the same encoder. Both the camera observations and subgoal images are first resized to 448x448 pixels.
For sampling history frames we use a stride of 1 second, 
and the entire history frames are dropped out entirely with probability 0.3. The rear view image (when available) is dropped out with probability 0.3 as well.

We employ a block-causal masking scheme, such that the observation tokens and the subgoal image tokens use bidirectional attention within themselves, and goal-image tokens can additionally attend the observations. The following text tokens use causal attention
(see attention mask visualization in the appendix). We also feed the proprioceptive state $\bq_t$ (including the history states) of the robot into the model backbone. Unlike \Pizs{} that uses discretized text tokens to represent $\bq_t$, \MethodName{} follows \texttt{MEM} and embeds the state using a linear projection that maps the state dimension to the backbone dimension. Each history state is treated as an individual token; if the history frame is dropped out, the corresponding state token is masked out as well.

The more lightweight ``action expert'' is a 860M-parameter transformer that is trained to predict continuous actions using flow matching objective. We use adaptive RMSNorm to inject timestep information for flow matching. The number of action tokens processed by the action expert is fixed at 50, representing an action chunk of 50 steps. The 50 tokens attend bidirectionally to each other and can also attend to the VLM backbone activations.

\MethodName{} also employs the training-time version of real-time action chunking (RTC)~\cite{black2025real,black2025ttrtc} for generating smooth action trajectories in the presence of inference delay. During training, we simulate delays of 0 to 12 timesteps, corresponding to a maximum inference latency of 240ms on a 50Hz robot.

\subsection{Training with subgoal images} 
When training \MethodName{} to handle subgoal images, we need the model to accommodate goals with different delays and different levels of image quality, including images generated by our world model. This requires carefully selecting which subgoals are provided as context to the model when training. We train on a combination of real images from future timesteps of the training trajectory and generated images. We found the following sampling scheme to be effective for selecting the timesteps for the real images:
with probability $0.25$, we sample the end-of-segment images (consistent with the prediction target for the world model), and with probability $0.75$ we sample future images uniformly from 0--4 seconds ahead of the current timestep. In addition to these real images, we mitigate the train-test mismatch between real and generated images by also sampling a large number of subgoal images from the world model, and constructing additional training examples with these \emph{generated} images added into the context of \MethodName{} instead of the real future images.

\section{Prompting \texorpdfstring{\MethodName{}}{pi0.7} at runtime}
\label{sec:runtime}

At runtime, we configure \MethodName{} to run with different forms of context depending on the desired behavior \textbf{without any task-specific post-training}. For any task we always prompt the model with the control mode and episode metadata. 
For choosing the episode metadata, we follow
\begin{itemize}[leftmargin=10pt]
    \item Overall speed: set per-task to the $15^\text{th}$ percentile of the episode length from the task.
    \item Overall quality: always set to 5, which is the highest score.
    \item Mistake: always set to false, meaning no mistake.
\end{itemize}
The subtask instruction $\rawtext_t$ is provided either by a learned high-level language policy or by a human supervisor for coaching (see Sec.~\ref{subsec:hl}).
When the subgoal images are used, we refresh the subgoal images whenever the semantic intent changes (i.e., new $\rawtext_t$), or after $\Delta=4$ seconds have elapsed since the last subgoal image was produced, whichever happens first. See Algorithm~\ref{alg:testtime} for the full workflow. We apply asynchronous inference: the visual subgoal and subtask instruction generation happens in separate threads and the VLA inference always uses the latest ones available.

For all experiments we use 5 denoising steps to generate the 50-step action chunks and execute $\hat H \in \{15, 25\}$ steps out of the chunk. Since each prompt component is trained with dropout, \MethodName{} can also be used with classifier-free guidance (CFG) \cite{ho2022cfg} for any part of the prompt, for example to guide the generated actions toward higher speeds. Concretely, each action denoising step follows
\begin{equation*}
\begin{aligned}
    \nabla_\ba &\log \pi_\theta(\ba_{t:t+H} \vert \bo_t, \mathcal{C}_t) + \\ 
    \beta &( \nabla_\ba \log \pi_\theta(\ba_{t:t+H} \vert \bo_t, \mathcal{C}_t) - \nabla_\ba \log \pi_\theta(\ba_{t:t+H} \vert \bo_t, \mathcal{C}_t^\text{uncond})),
\end{aligned}
\end{equation*}
where $\mathcal{C}_t^\text{uncond}$ denotes the set of context used in ``unconditional'' mode and $\beta$ is the CFG weight. While any part of the context could be dropped out, we apply CFG on the episode metadata to elicit strong performance in dexterous tasks. We use moderate values of $\beta \in \{1.3, 1.7, 2.2\}$.

\begin{algorithm}[t]
\caption{Prompting \MethodName{} at test time}
\label{alg:testtime}
\begin{algorithmic}[1]
\State \textbf{Input:} initial observation $\bo_0$, task instruction $\lang$, episode metadata $m$, control mode $c$
\State Initialize subtask $\rawtext$ (from high-level policy or coaching)
\State $\bg^{\star} \sim p_{\psi}(\bg^{\star} \mid \bo_0, \rawtext, m)$
\State $\mathcal{C} = \{\lang, \rawtext, \bg^{\star}, m, c\}$
\State $\ba_{t:t+H} \sim \pi_{\theta}(\ba \mid \bo_{t-T:t}, \mathcal{C})$ \Comment{Optional: CFG}
\For{$t=0,1,2,\ldots$}
    \If {$\rawtext$ changed \textbf{or} $\Delta$-second timer elapsed}
        \State $\bg^{\star} \sim p_{\psi}(\bg^{\star} \mid \bo_t, \rawtext, m)$ \Comment{Non-blocking (async)}
        \State $\mathcal{C} = \{\lang, \rawtext, \bg^{\star}, m, c\}$
    \EndIf
    \If {$\hat H$ steps elapsed since last inference}
        \State $\ba_{t:t+H} \sim \pi_{\theta}(\ba \mid \bo_{t-T:t}, \mathcal{C},\ba_{t:})$ \Comment{Async w/ RTC}
    \EndIf
    \State Execute $\ba_t$
\EndFor
\end{algorithmic}
\end{algorithm}

\section{Robot system details}
\label{sec:robots}
\begin{figure}[h]
    \centering
    \includegraphics[width=\linewidth]{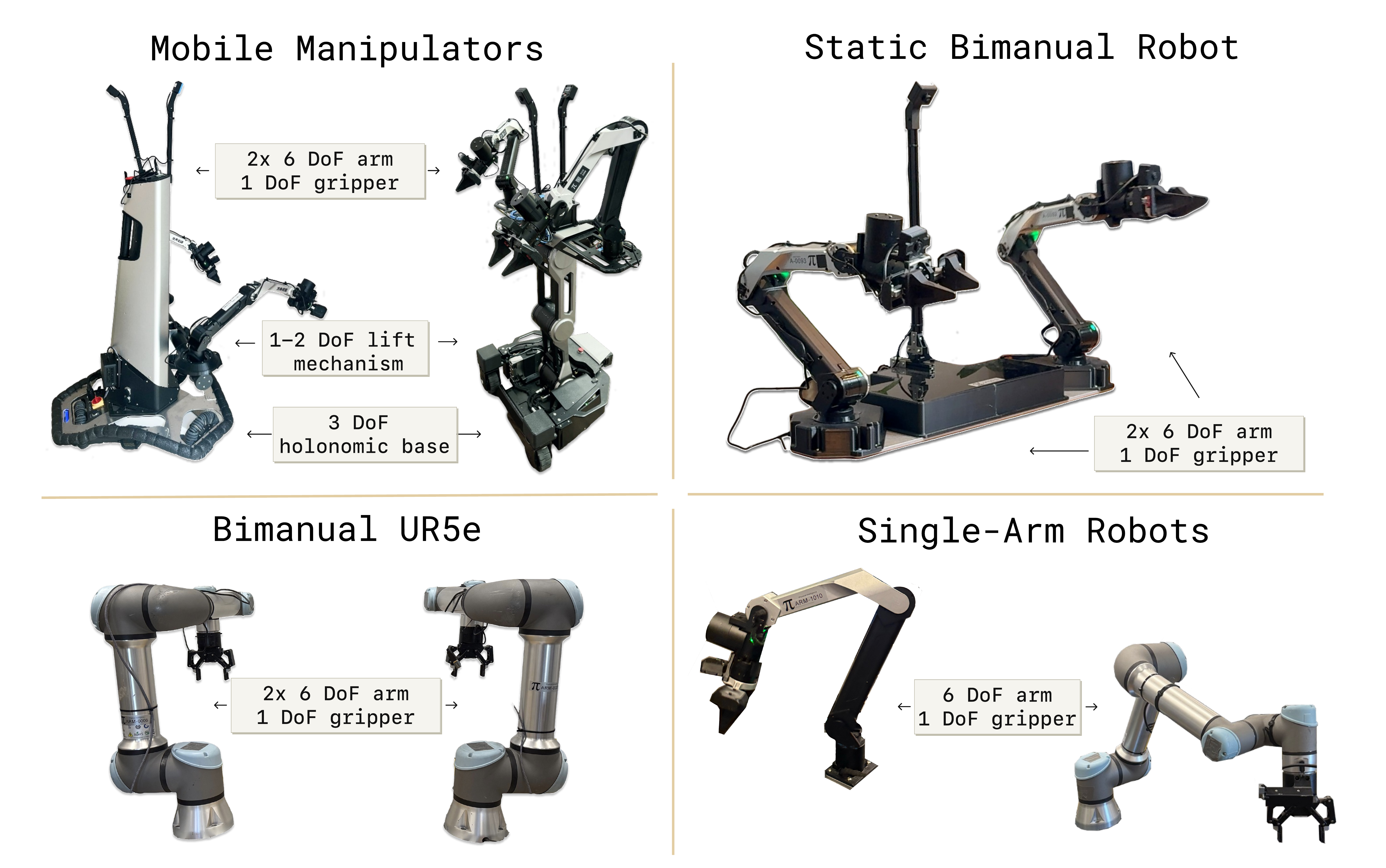}
    \caption{\textbf{Illustrations of some of the robots in our experiments.} We evaluate \MethodName{} on a variety of robots, including bimanual mobile manipulators (left), static bimanual robots (middle), and a bimanual UR5e setup (right) that we use for cross-embodiment experiments.}
    \label{fig:robots}
\vspace{-5pt}
\end{figure}

\begin{figure*}[h!]
    \centering
    \includegraphics[width=\linewidth]{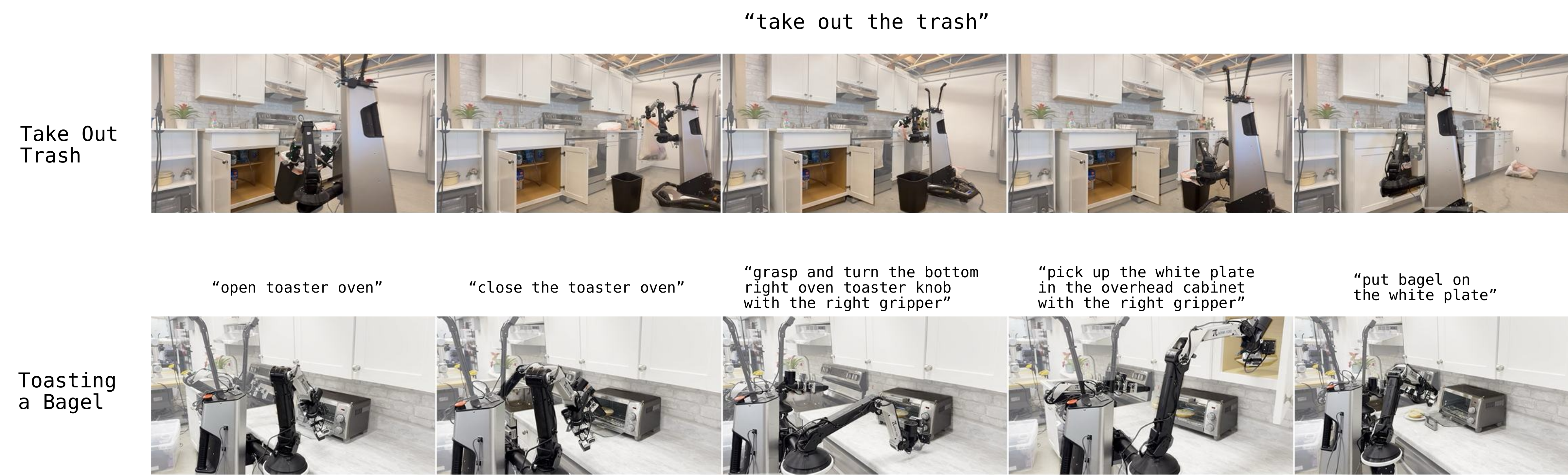}
    \caption{\textbf{Illustration of selected evaluation tasks.} We evaluate \MethodName{} on a number of tasks, and two of the more longer-horizon ones are visualized here. For some tasks such as ``Take Out Trash'', we provide a coarse instruction like ``take out the trash'' and \MethodName{} performs the full long-horizon task. For other tasks which do not appear in the training data for \MethodName{} such as ``Toasting a Bagel'', we can leverage the strong language following capabilities of \MethodName{} to coach it to perform the task with a series of detailed instructions that break down the task step-by-step.}
    \label{fig:multi_task_film_strip}
\vspace{-6pt}
\end{figure*}

We deploy \MethodName{} in a variety of robot platforms (Fig.~\ref{fig:robots}), including bimanual mobile manipulators with two 6 DoF arms, static bimanual manipulators with lightweight 6 DoF arms (``BiPi''), and a bimanual UR5e system with Robotiq grippers, which we use for cross-embodiment experiments. Additional generalization and language following experiments use a single-arm 6 DoF system, using the same arms as the BiPi platform. Note that while a large fraction of our data is collected with arms that resemble the BiPi platform, the UR5e arms that we use for cross-embodiment testing are significantly longer, have a different morphology, and are much heavier. In practice, the UR5e arms need to employ a different manipulation strategy due to the shape of the arms, their positioning over the table (on the sides rather than at one edge), and the shape of the gripper and fingers, making cross-embodiment transfer to this platform a significant challenge.
All manipulators use parallel-jaw grippers. The UR5e robots run at 20 Hz, while all other robots run at 50 Hz. Each robot has a front-facing camera as well as wrist camera on each arm, and the mobile robots also have a rear-facing camera.
The action output of the \MethodName{} model is applied on each robot using a simple PD controller. For commanding end-effector movement, we apply numerical inverse kinematics to convert target end-effector poses into target joint positions.

\begin{figure*}[h!]
    \centering
    \includegraphics[width=\textwidth]{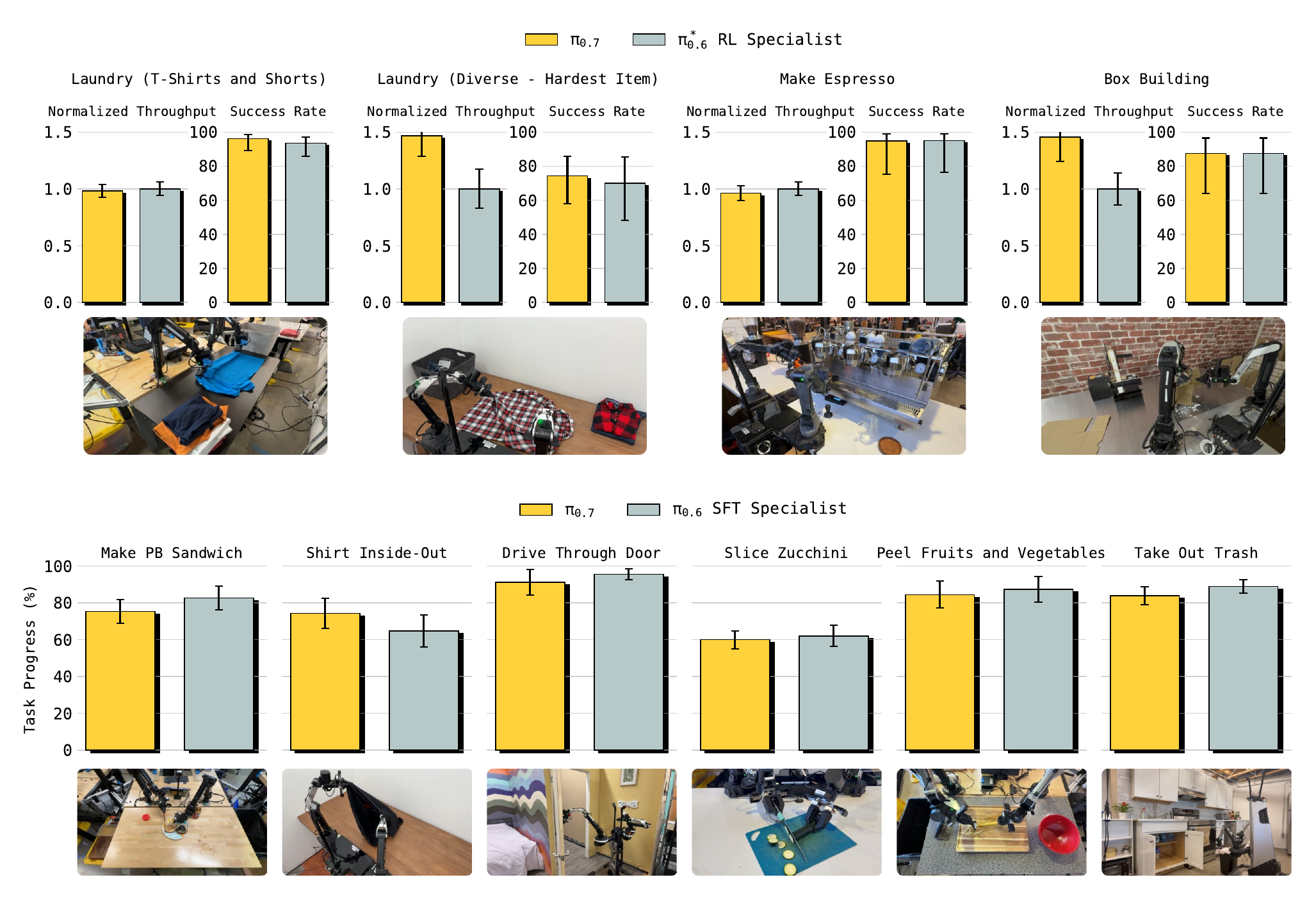}
    \caption{\textbf{Out-of-the-box dexterity: }\MethodName{} can perform a wide range of highly dexterous tasks directly out of the box. We consider tasks from \Piszs{} \citep{pistar06} (top row) and a number of other dexterous tasks including ones from the ``Robot Olympics'' experiments (bottom row). For the tasks from \Piszs{}, we report success rate and normalized throughput (relative to the specialist model; raw throughput means successes per hour), while for other tasks we report task progress. We find that the same \MethodName{} model can match the performance of the task-specific post-trained specialist policy from \Piszs{} or \Pizs{} for each of these tasks, and even achieve higher throughput than the RL specialists in diverse laundry folding and box building.}
    \label{fig:distillation_results}
\vspace{-10pt}
\end{figure*}

\begin{figure*}[h!]
    \centering
    \includegraphics[width=\textwidth]{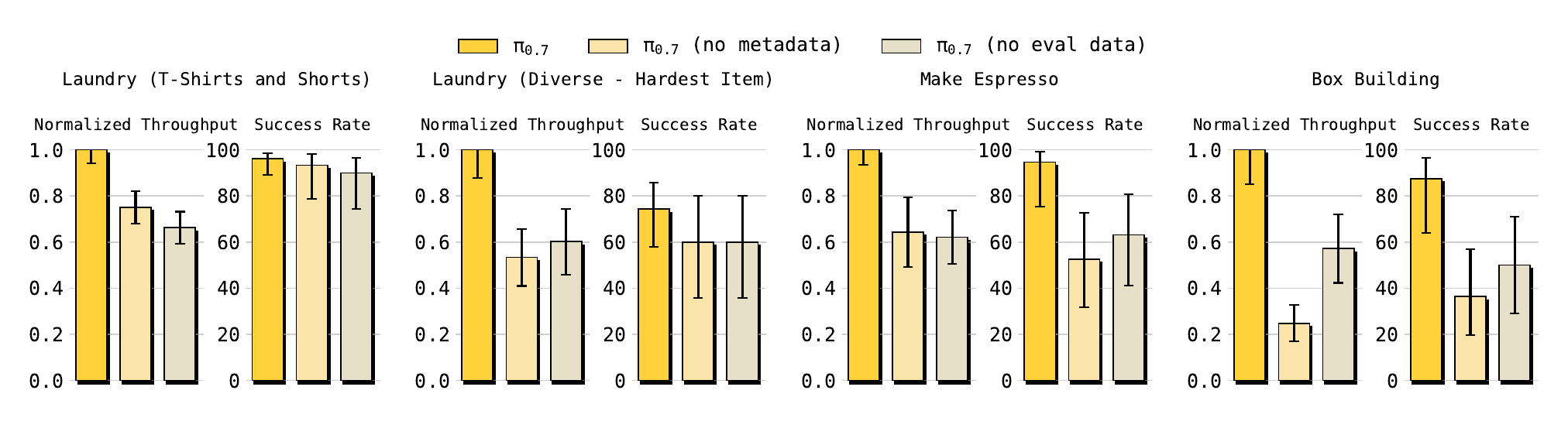}
    \caption{\textbf{Impact of prompt composition and evaluation data on out-of-the-box performance: }We compare \MethodName{} with two ablations: one that does not include episode metadata in the context, \MethodName{} (no metadata), and another that does not include data from autonomous evaluation episodes during training, \MethodName{} (no eval data). We find that \MethodName{} outperforms both \MethodName{} (no metadata) and  \MethodName{} (no eval data) across the board, with the gap most prominent in throughput. Throughput (successes/hour) here is normalized relative to \MethodName{}.}
    \label{fig:distillation_ablations}
\vspace{-5pt}
\end{figure*}

\begin{figure}[h]
    \centering
    \includegraphics[width=\linewidth]{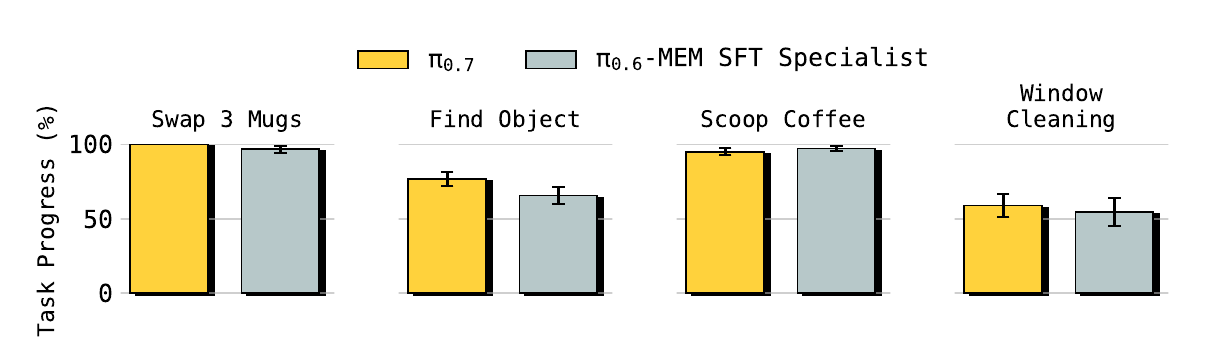}
    \caption{\textbf{Tasks that require memory: }\MethodName{} can also perform tasks that require explicitly keeping track of prior context, achieving similar or better performance compared to the specialist policies with memory fine-tuned to some of the tasks in the MEM paper~\citep{torne2026mem}.}
    \label{fig:memory}
\vspace{-10pt}
\end{figure}

\begin{figure*}[h]
    \centering
    \includegraphics[width=\textwidth]{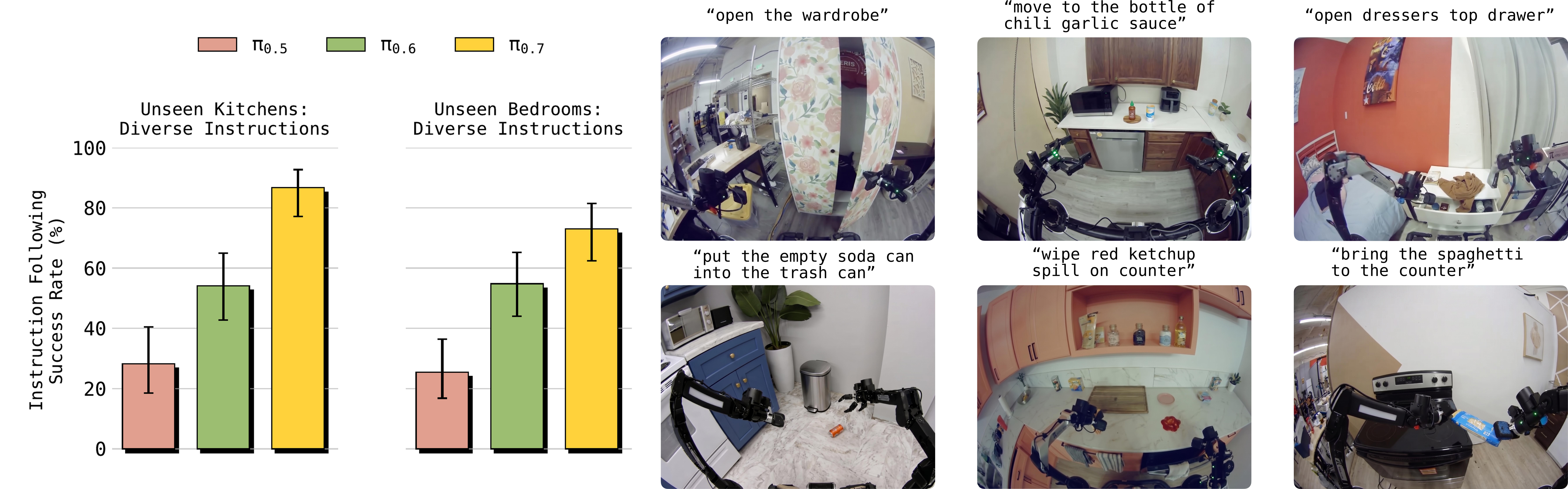}
    \caption{\textbf{Broad instruction following in novel environments: } We evaluate \MethodName{} on 14 instruction following scenarios, each of which involve following a sequence of 3-6 open-ended instructions, across 4 unseen kitchen and 2 unseen bedroom environments. We report the instruction following success rate, the percentage of total instructions that were correctly followed across all evaluations. We find that \MethodName{} significantly outperforms \Pizf{} and \Pizs{} across the board, attaining high absolute success rates.}
    \label{fig:instruction_following}
\vspace{-8pt}
\end{figure*}

\begin{figure}[h!]
    \centering
    \includegraphics[width=\linewidth]{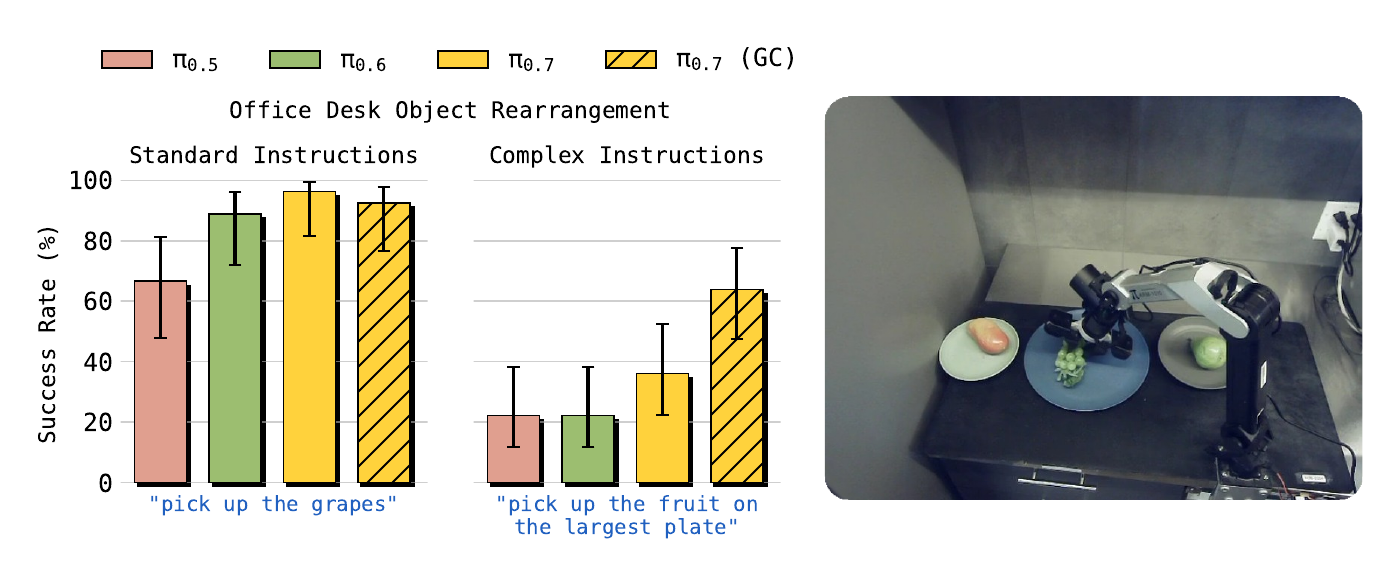}
    \caption{\textbf{Following complex referential instructions: }\MethodName{} and prior models all succeed on the simpler re-arrangement instructions (instructions include ``pick up the spoon'', ``put the spoon to the left of the fork'' and ``put the spoon to the right of the fork''), but \MethodName{} performs significantly better on the complex and unusual instructions (instructions include ``pick up the largest bowl on the table'', ``pick up the object I would use to eat soup'' and ``pick up the fruit on the largest plate''). Including subgoal images generated by a lightweight world model (\MethodName{} (GC)) further boosts instruction following performance, making \MethodName{} significantly more capable at following complex instructions.}\label{fig:instruction_generalization}
\vspace{-3pt}
\end{figure}

\begin{figure}[h!]
    \centering
    \includegraphics[width=\linewidth]{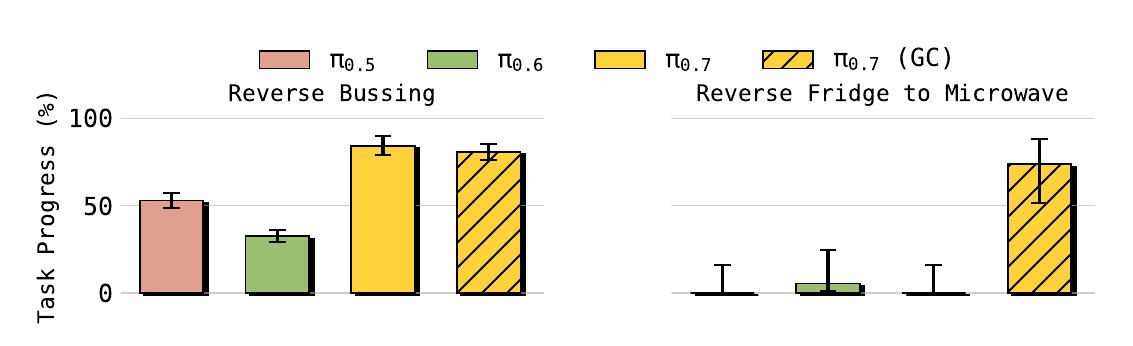}
    \caption{\textbf{Breaking dataset biases by following instructions: }the improved language-following performance of \MethodName{} enables it to break strong dataset biases. Prior models struggle on these data bias challenge tasks, which require following instructions that contradict the pattern in the data (e.g., putting dishes in the trash and trash in the dish bin). \MethodName, however, can follow instructions sufficiently well that it can break these strong biases to still perform the task. 
    Notably, for the ``Reverse Fridge to Microwave'' task, including subgoal images from the world model in the context (\MethodName{}(GC)) is critical for success.}
    \label{fig:compositional_generalization}
\vspace{-10pt}
\end{figure}

\section{Experimental Evaluation}
\label{sec:experiments}

In our experiments, we evaluate how well \MethodName{} can leverage diverse data sources to enable strong out-of-the-box performance, broad generalization, and more effective transfer, leveraging a variety of context modalities. Specifically, we study how well \MethodName{} can perform complex and dexterous tasks out of the box, particularly in comparison with more specialized RL-finetuned models (Sec.~\ref{sec:dexterity}), evaluate its ability to flexibly follow instructions to do a variety of different tasks (Sec.~\ref{sec:instruction_following}), study its transfer capabilities across embodiments (Sec,~\ref{sec:cross_emb_transfer}), and test its ability to compose skills in previously unseen ways to do new tasks (Sec.~\ref{sec:compositional_generalization}). Finally, we perform controlled experiments to study how the performance of \MethodName{} scales with increased task and context diversity in our robot datasets (Sec.~\ref{sec:scaling}).

\subsection{Out-of-the-box performance on challenging tasks}
\label{sec:dexterity}

\noindent \textbf{\MethodName{} achieves high performance on dexterous tasks without task-specific post-training.}
In our first set of experiments, we study how well \MethodName{} can master dexterous tasks that were seen in the training data, but where the goal is to perform these tasks as robustly and efficiently as possible. This is surprisingly difficult for prior robotic foundation models: often the best-performing policies are fine-tuned for specific downstream tasks, even if they use generalist pre-training~\citep{pistar06, pi06model}. We aim to answer: can the general-purpose \MethodName{} model match the performance of task-specific fine-tuned models on a variety of dexterous manipulation tasks?

We use the tasks shown in Fig.~\ref{fig:distillation_results}. These include the espresso making, box building, and laundry folding tasks that we previously used to evaluate the RL-trained \Piszs{} models~\citep{pistar06}, where we can directly compare the speed and robustness of the single general-purpose \MethodName{} model to the individual RL-finetuned specialist \Piszs{} models. We also study a number of other dexterous tasks, including some tasks from our previous ``Robot Olympics'' experiments (making a peanut butter sandwich, turning a shirt inside-out, and driving through a door) and a number of additional dexterous tasks, such as fully slicing up a zucchini, peeling a few fruits and vegetables (zucchini, cucumbers, and carrots), and a long-horizon task that involves replacing a trash bag in a trash can. We find that \MethodName{} achieves performance that is competitive with the RL specialists used in the \Piszs{} release \citep{pistar06} for all of the tasks considered in the paper directly out of the box (Fig.~\ref{fig:distillation_results}, first row), and even outperform the specialists in throughput in the difficult laundry and box building tasks. Additionally, we compare \MethodName{} to SFT specialists trained on top of \Pizs{} for a number of other dexterous tasks, and find that \MethodName{} is again able to closely match the performance of all specialist policies (Fig.~\ref{fig:distillation_results}, second row).

To understand how the training recipe of \MethodName{} affects performance, we additionally compare \MethodName{} with two ablations on the tasks from the \Piszs{} release: \MethodName{} (no eval data), which holds out all autonomous evaluation episodes from training (and thus cannot benefit from distilling agent rollouts from strong, e.g. RL trained, policies), and \MethodName{} (no metadata), which omits episode metadata from the context as shown in Fig.~\ref{fig:distillation_ablations} for the tasks used in the \Piszs~ release. Results suggest that \MethodName{} significantly outperforms both \MethodName{} (no eval data) and \MethodName{} (no metadata) on all tasks. Since policy evaluation data can vary widely in quality, training on this data, combined with rich metadata to disambiguate high and low quality behaviors, is critical for \MethodName{}'s strong performance on all of these challenging tasks.

\noindent \textbf{\MethodName{} achieves high performance on tasks that require memory without fine-tuning.}
In these experiments, we study how well \MethodName{} can perform tasks that require explicitly keeping track of previous observations~\citep{torne2026mem}. 
We compare the same single \MethodName{} model out of the box to the task-specific fine-tuned versions of \Pizs{} with memory used in~\citet{torne2026mem}, and find that \MethodName{} can achieve similar or better performance to the fine-tuned specialists on all of these tasks (Fig.~\ref{fig:memory}).

\subsection{Instruction following}
\label{sec:instruction_following}

In the next set of experiments, we study how well \MethodName{} can follow language instructions, including its ability to carry out a variety of different contexts and follow referential instructions that differ systematically from the training data. Our experiments focus specifically on performing tasks in messy environments where there are many possible tasks that the robot could perform, requiring \MethodName{} to pay careful attention to the provided instruction in order to succeed. We find that \MethodName{} displays instruction following capabilities that significantly improve upon our prior models \Pizf{}~\cite{black2025pi05} and \Pizs~\cite{pi06model}.

\noindent \textbf{\MethodName{} can be flexibly prompted to perform a wide variety of different tasks.} Language following has presented a notorious challenge for robotic foundation models, particularly with open-vocabulary instructions that do not correspond directly to ones seen in training. In these experiments, we aim to study the breadth of \MethodName{}'s capabilities, to answer: can \MethodName{} better handle a wider variety of language instructions than prior models?

We evaluate \MethodName{} on a diverse set of instructions across 4 unseen kitchens and 2 unseen bedrooms that were not present in the training data (Fig.~\ref{fig:instruction_following}). Each experiment tests whether the robot can follow a 3 to 6 step sequence of instructions to achieve a specific goal. The tasks require a variety of different realistic tasks in kitchen and bedroom environments, including re-arranging and tidying items, interacting with furniture, and cleaning up spills. This combination of new test environments and diverse instructions presents a major challenge for robotic foundation models, which can struggle to follow even simple instructions in seen environments. We find that \MethodName{} is able to significantly outperform \Pizf{} and \Pizs{}, with a high overall instruction following success rate.

\noindent \textbf{\MethodName{} can handle out-of-distribution referential instructions. }
Because of the breadth and diversity of our training data, it is generally difficult for us to quantify how \emph{novel} the test instructions are. In the next set of experiments, we intentionally designed a set of instructions that are unusual, refer to objects in unconventional ways, or require understanding spatial relations. In Fig.~\ref{fig:instruction_generalization}, we compare \MethodName{} and prior models on a set of object re-arrangement instructions, broken up into \emph{standard} and \emph{complex} instructions. The standard instructions are phrased in a similar way to language instructions in the training data. The complex instructions use unusual language or complex spatial references, such as ``pick up an object I would use to eat soup" or ``pick up the fruit on the largest plate." In these experiments, we also evaluate \MethodName{} with and without subgoal images, which are generated by a lightweight world model as discussed previously. We can see that \MethodName{} improves over prior models on the complex instructions, and the use of subgoal images further boosts its performance, importing semantic understanding from the world model.

\begin{figure*}[h!]
    \includegraphics[width=\textwidth]{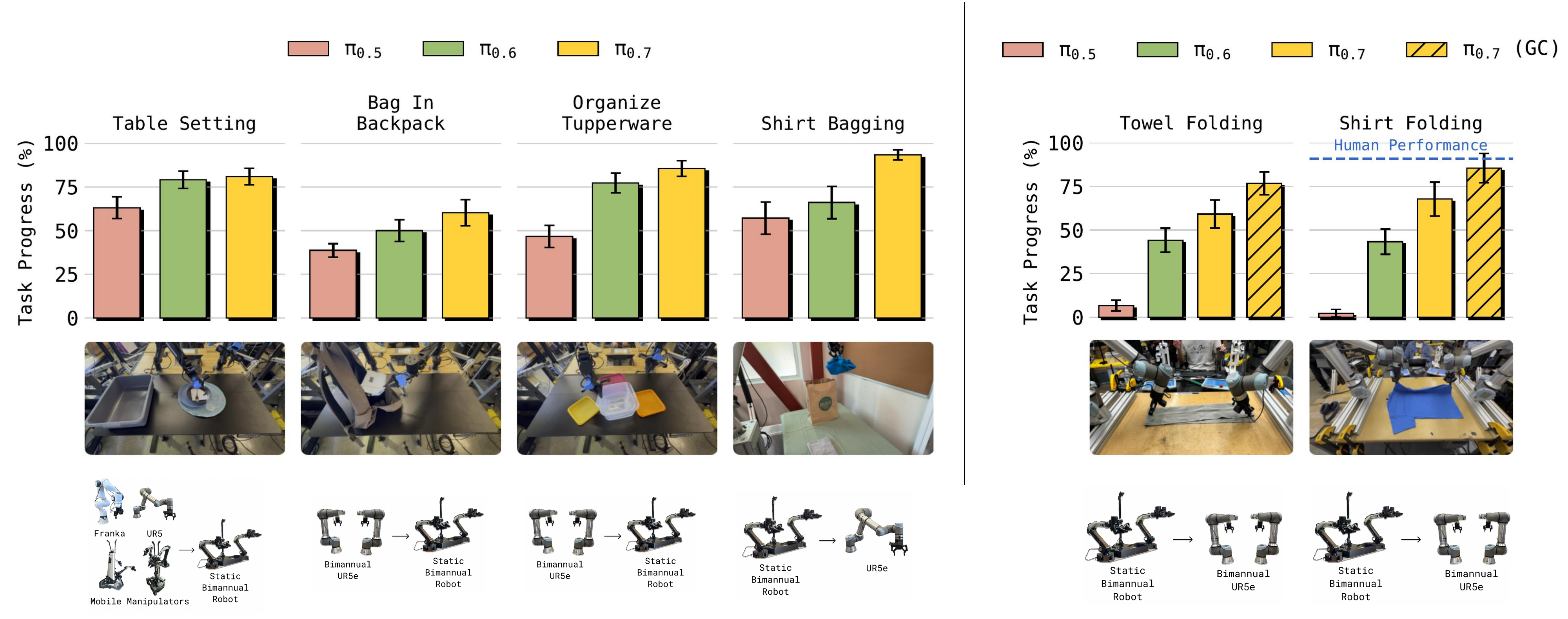}
    \caption{\textbf{Cross-embodiment transfer: }Left: Both \MethodName{} and prior models achieve strong cross-embodiment transfer directly out of the box on simpler re-arrangement or repositioning style tasks. For example, for the ``Table Setting'' task, data was collected with a variety of different robots, and the task was tested on the static bimanual robot. All of the models performed well. For tasks that required transfer from the bimanual UR5e robot to the smaller static bimanual robot (``Bag In Backpack'' and ``Organize Tupperware''), the embodiment gap is larger, as there is only one source robot that is bigger and heavier than the target robot. Here, \Pizf{} performed very poorly, but \Pizs{} still performs quite well. Transferring from the smaller static bimanual robot to the UR5e for the ``Shirt Bagging'' task introduces the largest embodiment gap, and here the \MethodName{} model significantly outperforms prior models. Right: for the more dexterous tasks that require folding towels and t-shirts, the embodiment gap poses an even greater challenge. Data for these tasks was collected with the smaller static bimanual robot, and the task was evaluated on the larger bimanual UR5e platform. \MethodName{} was able to transfer this task successfully, and improved even more when using visual subgoal images generated by our lightweight world model. In fact, task progress matches the 
    ``zero-shot'' performance of our most experienced human teleoperators, who have operated a diverse set of robots and attempted this task on the UR5e for the first time.}
    \label{fig:cross_embodiment}
\vspace{-5pt}
\end{figure*}

\noindent \textbf{\MethodName{} can follow instructions that go against dataset biases. }
Dataset bias presents a major challenge for instruction following: if the robot always does the same thing in a given scene, a model trained on such data will often ignore language in these scenes, blindly copying the behavior seen in the data. In the next experiment, we construct scenarios that suffer from this problem, and test if we can prompt \MethodName{} to go \emph{against} the natural bias in the dataset.
We constructed two tasks: ``Reverse Bussing'' and ``Reverse Fridge to Microwave.'' In our dataset, the ``bussing'' task involves putting trash in a trash bin and dishes in a bussing bin. For the ``Reverse Bussing'' task, the robot is asked to do the opposite: put the trash in the bussing bin, and the dishes in the trash. The ``Fridge to Microwave'' task requires taking food out of the fridge and putting it in the microwave, and we did not collect data going the other way. At test time, in the ``reverse'' version of this task (``Reverse Fridge to Microwave''), we prompt the robot to take the food from the microwave to the fridge, violating the bias in the dataset.

The results in Fig.~\ref{fig:compositional_generalization} show that \MethodName{} significantly improves over prior models on these tasks. This suggests that \MethodName{} has significantly better language following capabilities, and pays enough attention to the instructions to overcome the bias in the data for these tasks. On the ``Reverse Fridge to Microwave'' task, conditioning on generated subgoal images (\MethodName{} (GC)) is critical for success, as the world model can generate subgoals based on textual instructions effectively by leveraging web-scale image generation pre-training.

\subsection{Cross-embodiment transfer} \label{sec:cross_emb_transfer}

While a number of models have used data from multiple different robot embodiments~\citep{open_x_embodiment_rt_x_2023,yang2024pushing, team2024octo,wang2024hpt}, zero-shot transfer of complex tasks from a source embodiment to a target robot that has never seen the task presents a major challenge. In these experiments, we aim to study whether cross-embodiment transfer is an \emph{emergent} property of \MethodName{}. Namely, can \MethodName{} directly transfer capabilities to robot embodiments where no task-specific data was ever collected?

We find that for several tasks, \MethodName{} succeeds on target embodiments entirely out of the box, on tasks for which the target robot has no training data. For more modest embodiment differences, the \Pizf{} and \Pizs{} models also show some amount of emergent cross-embodiment transfer. However, as the gap in the robot morphology increases, effectively transferring complex skills requires more significant changes in strategy. In these cases, \MethodName{} significantly outperforms the prior models, and even matches the ``zero-shot'' performance of human teleoperators for the shirt folding task, as we discuss below. Experimental results below apply joint-space control, as we find that end-effector control does not yield noticeable gains in performance with our prior models (Appendix~\ref{app:xemb_joint_vs_ee}).

\noindent \textbf{Zero-shot cross-embodiment transfer for object re-arrangement tasks. }
We first study a set of simpler object rearrangement tasks, where we test \MethodName{} on a robot other than the one that was used to collect data for that task. The results are presented in Fig.~\ref{fig:cross_embodiment}. For the first task, ``Table Setting,'' the data was collected with several different robot types, including mobile, static, and single-arm systems. This is the most favorable setting for cross-embodiment transfer, since the models can infer the common structure of the task from multiple different robots. When evaluated on a static bimanual robot system, we find that all methods show strong signs of cross-embodiment transfer. However, when we increase the embodiment gap more significantly and test policies on the smaller static bimanual platform when all data was collected on the larger UR5e bimanual platform (``Bag In Backpack'' and ``Organize Tupperware'' tasks), we find that the performance of \Pizf{} degrades significantly, while both \Pizs{} and \MethodName{} still are able to achieve strong performance. We then increase the embodiment gap even more, studying transfer for a task where data was collected on the smaller static bimanual platform and evaluated on a \emph{single-arm} UR5e platform (``Shirt Bagging'' task). This necessitates a significantly different strategy, as the target robot has only one arm, is much larger, and also heavier. Here, \MethodName{} significantly outperforms the prior models.

\begin{figure}[h!]
    \centering
    \includegraphics[width=\linewidth]{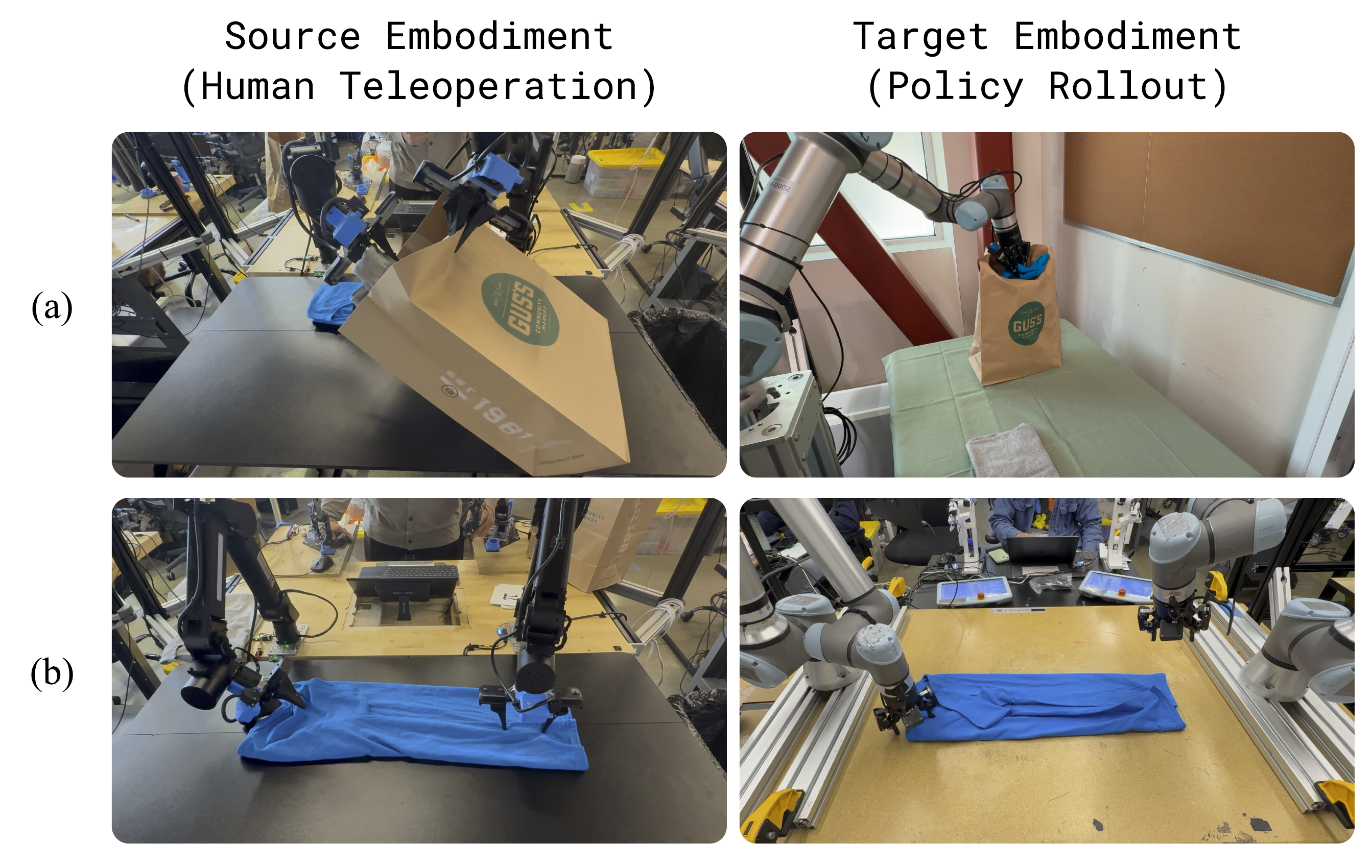}
    \caption{\textbf{Cross-embodiment transfer produces emergent strategies adapted to the target embodiment.} (a) On the source robot, human teleoperators use one arm to hold the bag open while the other performs insertion. On the UR5e target robot, \MethodName{} instead discovers a single-arm pick-and-place strategy suited to the robot's greater reach. (b) Human teleoperators approach the shirt with a tilted end-effector on the source robot, while \MethodName{} produces vertical grasps on the UR5e, which is more suitable for the larger robot's arm placement. In both cases, the policy goes beyond replicating source behavior, discovering manipulation strategies for the task that are better suited to the target embodiment.}
    \label{fig:human_vs_policy}
\vspace{-10pt}
\end{figure}

Notably, successful transfer often requires the policy to discover new manipulation strategies suited to the target morphology rather than simply replicate the source behavior. For example, the shorter static bimanual robot must use one arm to hold the bag open while the other performs insertion, whereas the taller UR5e arm can accomplish the same task with a single-arm pick-and-place (Fig.~\ref{fig:human_vs_policy} (a)). Despite this significant morphological gap, the model applies the appropriate strategy for each embodiment, demonstrating cross-embodiment transfer that goes beyond mimicking the original robot's motion.

\begin{figure*}[h]
    \centering
    \includegraphics[width=\textwidth]{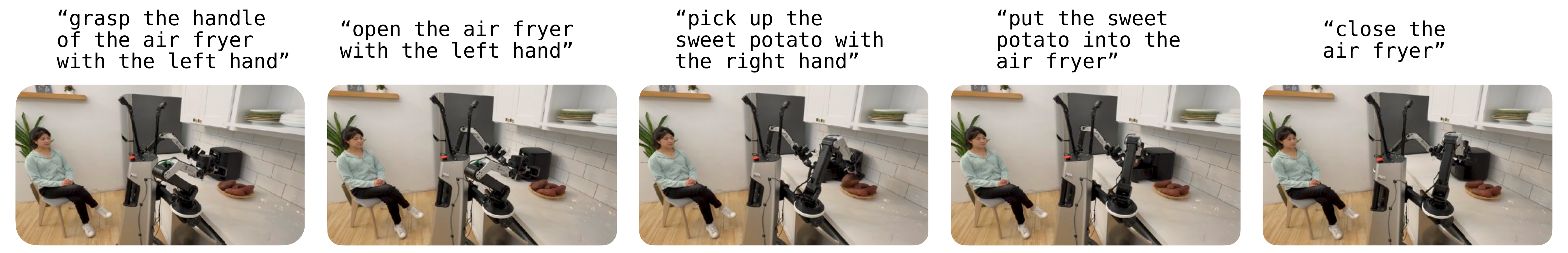}
    \caption{\textbf{Example of language coaching: } We can ``teach'' a new task to \MethodName{} by providing step-by-step verbal instructions. Because of its language following ability, \MethodName{} can perform new tasks successfully under user instruction, and these instructions can then by used to train a high-level policy that prompts \MethodName{} so that it can perform the task fully autonomously.}
\label{fig:air_fryer_coaching}
\vspace{-5pt}
\end{figure*}

\begin{figure}[h]
    \centering
    \includegraphics[width=\linewidth]{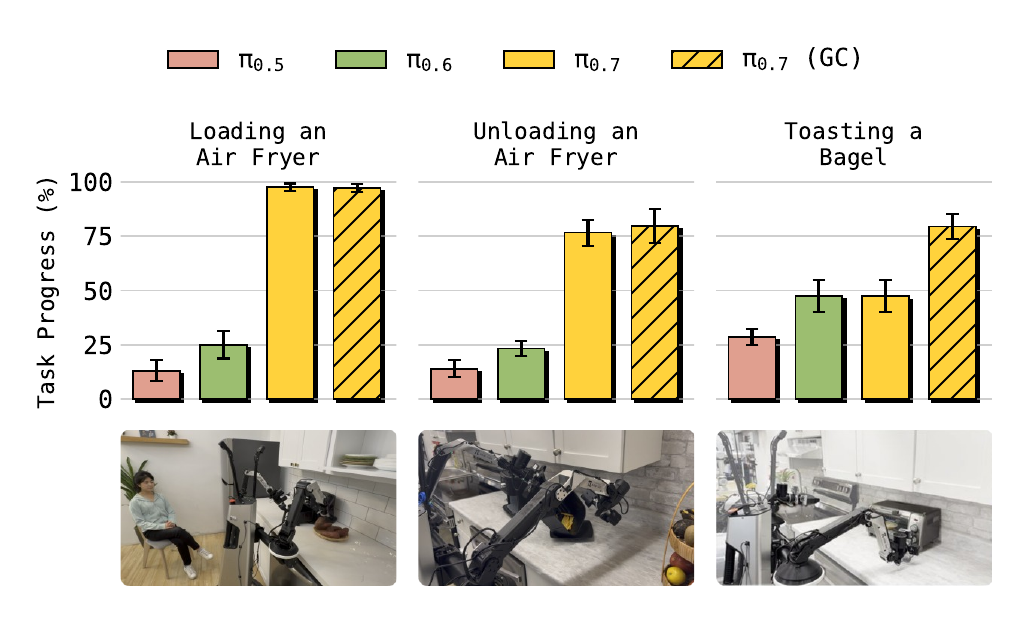}
    \caption{\textbf{Coaching to perform new long-horizon tasks: }Because \MethodName{} can follow language instructions effectively, even for unfamiliar skills, it can be ``coached" to perform a number of unseen, longer horizon tasks both when conditioned on language and generated subgoal images (\MethodName{} (GC)). Prior models generally lack the language following ability needed to follow the coaching commands, and thus achieve very poor performance.}
\label{fig:long_horizon_task_generalization}
\vspace{-5pt}
\end{figure}

\begin{figure}[h]
    \centering
    \includegraphics[width=\linewidth]{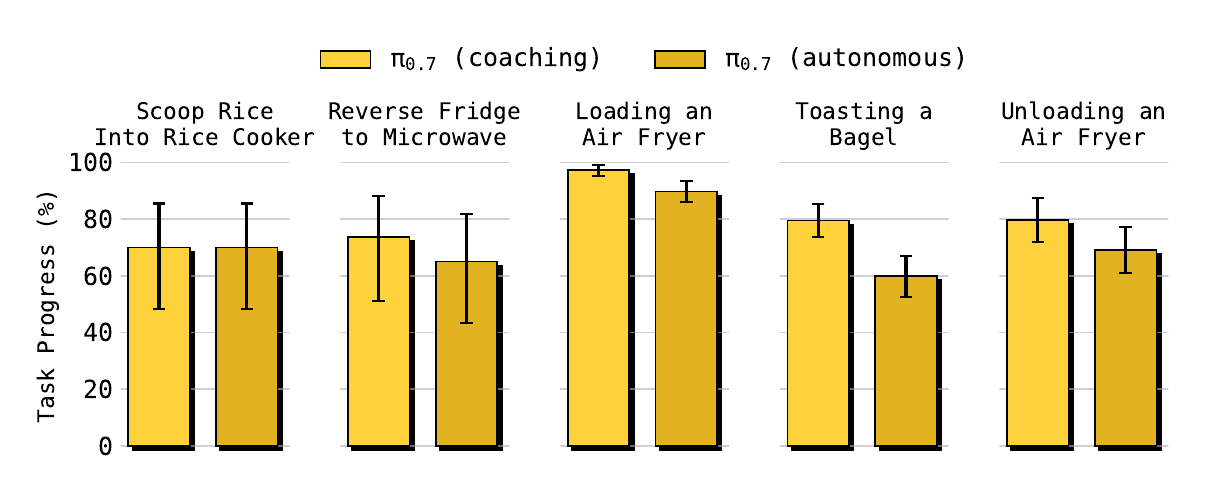}
    \caption{\textbf{Acquiring new autonomous capabilities with coaching: } We can use the coaching episodes collected for a number of different unseen tasks to train a high-level policy to automatically prompt \MethodName{} in accordance with the coaching episodes. This allows us to create fully autonomous policies for these tasks (\MethodName{} (autonomous)) that closely match the performance of the policy with live human coaching (\MethodName{} (coaching)) \emph{without collecting any additional data with teleoperation or any other kind of low-level actions}.}
    \label{fig:coaching}
\vspace{-5pt}
\end{figure}

\begin{figure}[h]
    \centering
    \includegraphics[width=\linewidth]{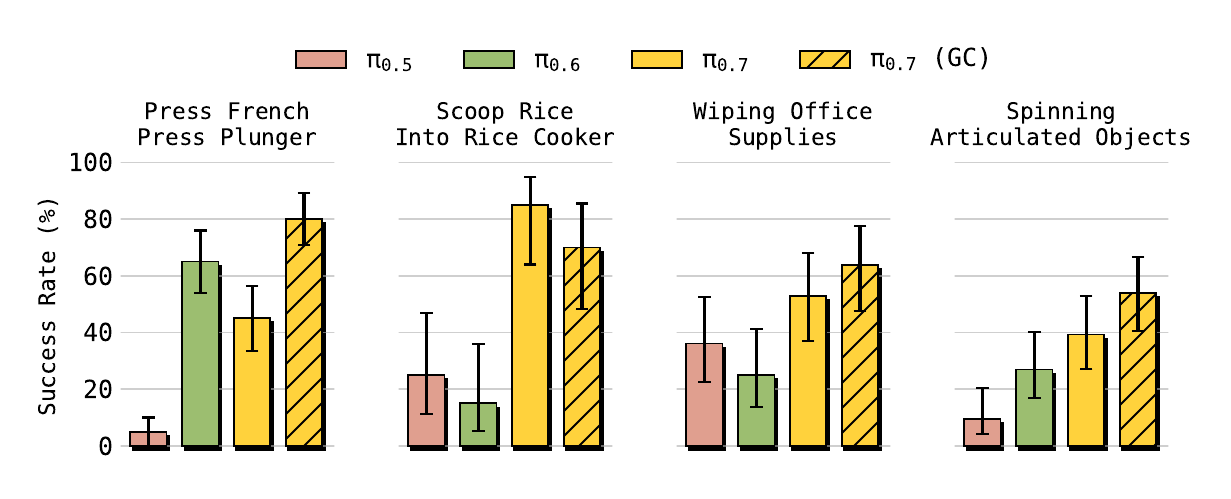}
    \caption{\textbf{Performing new short-horizon tasks: }\MethodName{} can perform a number of new short-horizon tasks directly out of the box, including scooping rice into a rice cooker, spinning various objects such as a gear set and desk fan, and wiping down objects with a cloth, such as a ruler and headphones, despite no data being collected for any of these tasks. \MethodName{} shows roughly equally strong performance when conditioned directly on language instructions (\MethodName{}) or generated image goals (\MethodName{} (GC)).}
    \label{fig:task_generalization}
\vspace{-10pt}
\end{figure}

\noindent \textbf{Zero-shot cross-embodiment transfer for dexterous tasks. }
Dexterous tasks, such as laundry folding, present a bigger challenge for cross-embodiment transfer. Such tasks require more precise manipulation skills than just grasping and repositioning objects. Successfully folding a t-shirt requires a sequence of precise grasps and placements, and the correct grasp angle might vary with the robot's reachable workspace and gripper orientation. Most of our folding data was collected with lightweight static bimanual robots (see Fig.~\ref{fig:robots}). We did not collect any laundry folding data with the bimanual UR5e system. The morphology of this system, its reachable workspace, and its dynamics (e.g., higher internal inertia) differ significantly from the robots with which we collected laundry folding data. We found the UR5e generally harder to teleoperate and less suitable for very precise grasps, suggesting that folding laundry with this robot requires a change in the manipulation strategy.

\begin{figure*}[h]
    \centering
    \includegraphics[width=\textwidth]{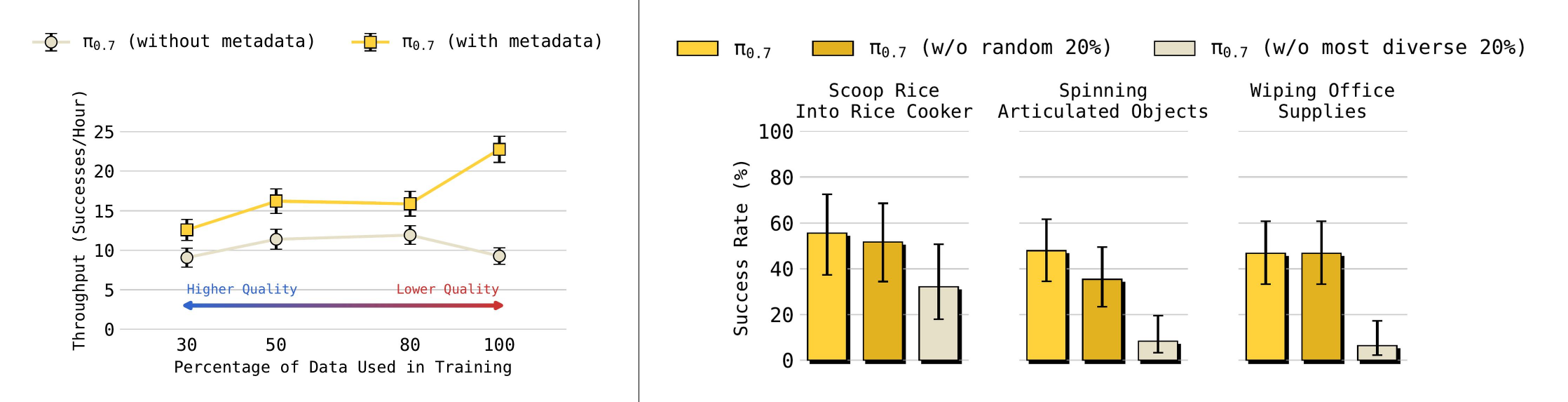}
    \caption{\textbf{Scaling of generalization performance with diverse context and data: } Left: We find that \MethodName{} (with metadata) can continuously improve its performance when it is trained on larger datasets, even when the average quality of the data actually decreases. By contrast, without training on rich conditioning information, we find that \MethodName{} (without metadata) actually can degrade in performance as more lower quality data is introduced. Right: When \MethodName{} is trained without our robot data with the highest task diversity, its performance degrades substantially. This suggests that \MethodName{} is able to utilize the task diversity in our robot data to drive substantial improvements in compositional task generalization.}
\label{fig:generalization_and_conditioning}
\vspace{-8pt}
\end{figure*}

\MethodName{} was able to successfully fold both towels and shirts on the bimanual UR5e system (Fig.~\ref{fig:cross_embodiment}, right).
On the source robot (the static bimanual robot in Fig.~\ref{fig:robots}), human operators often approach the cloth with a tilted end-effector to secure the fabric against the table before lifting. On the UR5e, \MethodName{} instead uses a vertical grasp that is better suited to the arm's kinematics --- a strategy that is different from the training data on the source robot but more suitable for the target embodiment (Fig.~\ref{fig:human_vs_policy} (b)). We also find that using subgoal image generation with our world model significantly improves performance (labeled \MethodName{} (GC) in the plots), as the world model can more effectively construct visual analogies between the source and target robots. The generated subgoals predict what kind of grasps and clothing configurations would be reasonable for the target robot, and the model incorporates this additional context as a hint for selecting better actions.

To contextualize these results, we conducted a human subject study with 10 experienced teleoperators (mean 375 hours of teleoperation experience across all robots, all within the top 2\% by experience). Like the policy, these operators had extensive experience on the source embodiment, but had never attempted shirt folding on the UR5e bimanual system, making this a zero-shot cross-embodiment transfer setting for both humans and the policy. The human operators achieved 90.9\% task progress and an 80.6\% success rate, while \MethodName{} achieved 85.6\% task progress and an 80\% success rate, demonstrating performance comparable to these expert operators. The strong cross-embodiment transfer performance highlighted by this comparison is scientifically exciting, and it has a practical implication: dexterous skills may be transferred \textit{from} lightweight, low-cost platforms that are easy to teleoperate \textit{to} high-payload industrial arms, where collecting human demonstration data is substantially more expensive and difficult. More details on the human study setup and results are provided in Appendix~\ref{app:human_study}.

\subsection{Compositional task generalization}
\label{sec:compositional_generalization}

In the next set of experiments, we study how well \MethodName{} can perform new tasks by compositionally generalizing over the skills seen in training. This has been viewed as a kind of ``grand challenge'' for robotic foundation models: while prior models have demonstrated generalization over semantic concepts (e.g., reaching for an object with an unseen textual label), performing new tasks has proven elusive.

We found that for some short-horizon tasks, \MethodName{} can work well completely out of the box, despite no data being collected explicitly for these tasks. These tasks involve manipulating unfamiliar objects in new ways, such as wiping down a pair of headphones with a cloth, or spinning a desk fan (Fig.~\ref{fig:task_generalization}). For longer-horizon, more complex tasks, we find that we can actually use the instruction following capabilities of \MethodName{} to ``coach'' it through the task with language. This provides an exciting new way to teach \MethodName{} new tasks without collecting any additional training data.

\noindent \textbf{\MethodName{} can perform new short-horizon tasks out of the box. }
We find that \MethodName{} can perform short-horizon tasks such as pressing the plunger on a french press, scooping rice into a rice cooker, wiping down common office objects, and spinning various articulated items directly out of the box, despite no robot data being collected specifically for any of these tasks (Fig.~\ref{fig:task_generalization}). This is quite exciting --- since \MethodName{} can flexibly compose a wide variety of skills, we can often get it to perform new, simple tasks by just prompting it to get the desired behaviors.

\noindent \textbf{\MethodName{} can be coached to perform new longer horizon tasks purely with language. }
While \MethodName{} can be directly prompted to perform new short-horizon tasks, simply asking the robot to perform an unseen long-horizon task like cooking a sweet potato would not work: although \MethodName{} exhibits a high level of out-of-the-box generalization, these tasks are simply too complex, requiring up to 5 minutes of interaction with multiple stages. However, \MethodName{}'s language following abilities provide us with an exciting new path to teach the model such tasks: instead of providing demonstration data for each complex skill that we want the robot to learn, we can instead ``coach'' the robot to perform the new task with language, a bit like how one might teach the task to a person (Fig.~\ref{fig:air_fryer_coaching}). We set up several realistic multi-stage kitchen tasks: (1) ``Loading an Air Fryer'': using an air fryer to cook a sweet potato; (2) ``Unloading an Air Fryer'': dumping items out of an air fryer, and (3) ``Toasting a Bagel'': using a toaster to toast up a bagel. In each case, our robot data did not contain any training episodes for this task, although similar appliances were seen in different contexts in human data and external datasets. However, a person can walk the robot through task with step-by-step instructions, such as ``pick up the sweet potato'' and ``open the air fryer.'' We present the results of coaching \MethodName{} and comparisons in Fig.~\ref{fig:long_horizon_task_generalization}. Critically, none of the models have any \emph{action-level} data of these particular tasks, and in the ``coaching'' episodes, the environment and task are entirely unseen. We find that \MethodName{} can be coached much more effectively than prior methods to perform all of these tasks, and can perform even more effectively when conditioned on generated subgoal images.

\noindent \textbf{Coaching data can endow \MethodName{} with new capabilities. }
Since \MethodName{} can be coached to perform new tasks, we can actually use the step-by-step instructions in the coaching data to train a high-level language policy which can prompt \MethodName{} with the appropriate language instructions as it performs the task. This gives \MethodName{} the ability to perform fully unseen, long-horizon tasks without collecting any additional teleoperation data. We show the results of this experiment in Fig.~\ref{fig:coaching} for five different tasks. For all of these tasks, we find that we can successfully train an autonomous policy (\MethodName{} (autonomous)) that can roughly match the performance of the coaching episodes (\MethodName{} (coaching)) that were collected by prompting the model to perform the task.

\subsection{Can \texorpdfstring{\MethodName{}}{pi0.7} learn effectively from diverse and mixed-quality data?}
\label{sec:scaling}
In our last set of experiments, we perform a set of controlled ablation studies to understand whether \MethodName{} can effectively leverage large and diverse datasets, and whether its performance improves with dataset diversity. These questions are difficult to answer definitively, since the performance of such a model depends on a large number of factors, and very large datasets are very difficult to slice cleanly so as to ablate ``diversity". To provide some understanding of these questions, we first study whether \MethodName{} continues to improve on seen tasks when trained on increasingly larger but more mixed-quality datasets. Then, we study whether \MethodName{} can leverage datasets with high task diversity to drive improvements in generalization.

\noindent \textbf{\MethodName{} can effectively learn from mixed-quality data: }
Effectively learning from diverse robot data has so far been a significant challenge in training robotic policies. Designers will often carefully filter or curate the data to pull high-quality datasets with consistent strategies \cite{hejna2025robot, li2025gr}. However, data filtering is laborious, task-specific, and ends up throwing away a lot of valuable information. In these experiments, we aim to answer: can \MethodName{} learn more from data with diverse manipulation strategies?

To study this question, we consider the ``Laundry (T-Shirts and Shorts)'' task that we studied in Fig.~\ref{fig:distillation_results}. We annotated the data collected by human operators based on fold quality and speed, and split our dataset into 4 buckets consisting of (1) the top 30\% by quality and speed, (2) the top 50\%, (3) the top 80\%, and (4) all of the data. We then train new \MethodName{} models from scratch for data from each of the four buckets, using episode metadata or without it (8 models in total).
We find that while \MethodName{} (without metadata) can actually get worse when trained on larger, mixed-quality datasets, \MethodName{} (with metadata) is able to continuously improve as we train on more data, even though the dataset size increases corresponds to a decrease in average data quality (Fig.~\ref{fig:generalization_and_conditioning}, left). This suggests that our diverse prompting method effectively makes the model design \emph{more scalable}, in the sense that it benefits more from larger datasets, even when these larger datasets actually consist of lower quality data that harm models trained in the usual way. Episode metadata effectively disambiguates the different data quality and strategies within the large-scale datasets during \MethodName{} training, and enables prompting for the desired mode of behavior at test time.

\noindent \textbf{Can \MethodName{} translate increased dataset diversity into better generalization performance?}

To study this question, we compare \MethodName{} with the following ablations on some of the short-horizon unseen tasks from Fig.~\ref{fig:task_generalization}.
\begin{itemize}
    \item \MethodName{} (w/o most diverse 20\%): \MethodName{} but with the 20\% of our data with the highest task diversity removed.
    \item \MethodName{} (w/o random 20\%): \MethodName{} with a randomly sampled 20\% of our data removed to serve as a data-controlled comparison to \MethodName{} (w/o most diverse 20\%)
\end{itemize}

The comparison between \MethodName{} (w/o most diverse 20\%) and \MethodName{} (w/o random 20\%) allows us to understand the impact of data with high task diversity in a controlled way, as both models are trained on the same quantity of data. We find that across all tasks, \MethodName{} and \MethodName{} (w/o random 20\%) significantly outperform \MethodName{} (w/o most diverse 20\%), demonstrating that \MethodName{} is able to effectively ingest data with high task diversity and translate that data into performance improvements on short-horizon, unseen tasks (Fig.~\ref{fig:generalization_and_conditioning}, right).

\section{Discussion}

We presented \MethodName{}, a general-purpose robot foundation model that exhibits out-of-the-box compositional generalization, effective language following, and task performance that is competitive with more specialized models that are fine-tuned to individual dexterous tasks. At the core of \MethodName{} is a diverse prompting strategy, inspired by prompt expansion, where additional information about the episode is provided to the model during training and, optionally, at test time. This additional information includes more detailed language, episode metadata, and subgoal images. Our experiments show that by using diverse prompting and a larger and more diverse dataset, \MethodName{} can represent policies of many different qualities, and distill specialist performance back into one pre-trained model. \MethodName{}  acquires a number of emergent capabilities, such as the ability to transfer skills across robots, follow complex language commands, and generalize compositionally, recombining skills in new ways to solve new tasks.

While \MethodName{} generalizes broadly, the success rate for zero-shot generalization is (unsurprisingly) lower than in-distribution tasks: while seen tasks often have success rates in excess of 90\%, unseen tasks or unseen task-robot combinations have success rates in the 60-80\% range. An exciting direction for future work is to leverage the high steerability of \MethodName{} to efficiently learn from data in the test task, for example with more detailed language coaching or even with autonomous reinforcement learning.

A perhaps surprising limitation of our experiments is that it's practically difficult when training on such large and diverse datasets to definitively determine which tasks are truly ``seen'' or ``unseen'': while some of our generalization experiments (e.g., those in Sec.~\ref{sec:compositional_generalization}) use tasks for which we did not deliberately collect data, our dataset contains so many different scenes and behaviors that potentially related skills may well be present elsewhere in the data, either with a different label, or incidentally as part of performing other tasks. In many ways this mirrors the challenge of understanding generalization with large language models: determining what is truly novel becomes difficult, and the model may well be achieving generalization primarily by ``remixing'' skills and behaviors from other situations. However, we would posit that this in fact is the essence of \emph{compositional generalization}. Practically, whether the behaviors are truly new or merely novel combinations of seen parts, the ramifications are similar: instead of deliberating collecting targeted data for each new task that we want the robot to solve, a model that provides compositional generalization would allow the user to simply \emph{prompt} it to do the desired task. Models that can enable such compositional generalization at scale would transform how we approach robotic learning, making it possible to prompt, coach, and explain things to a robot rather than needing to collect additional action data.

\section*{Acknowledgements}

We thank our robot operators for data collection, evaluations, logistics, and video recording, and our technicians for robot maintenance and repair. See Appendix~\ref{app:contributions} for a full contributions statement.

\bibliographystyle{unsrtnat}
\bibliography{references}

\clearpage

\appendix

\section{}

\subsection{Contributions}
\label{app:contributions}

\noindent\textbf{Data collection and operations}. Ashwin Balakrishna, George Bokinsky, Thomas Charbonnier, Grace Connors, Michael Equi, Chelsea Finn, Lachlan Groom, Hunter Hancock, Karol Hausman, Connor Jacobsen, Rowan Jen, Marinda Lamb, Vishnu Mano, Nandan Marwaha, Aikys Mongush, Tyler Patterson, Charvi Sharma, Lucy Xiaoyang Shi, Laura Smith, Will Stoeckle, Anna Walling, Jason Wang, Samuel Whitmore, Blake Williams.

\noindent\textbf{Annotation and supplemental data.}. Ashwin Balakrishna, Karan Dhabalia, Danny Driess, Chelsea Finn, Haroun Habeeb, Rowan Jen, Chandra Kuchi, Karl Pertsch, Lucy Xiaoyang Shi, Will Stoeckle, Quan Vuong.

\noindent\textbf{Policy training and research.}. Bo Ai, Ashwin Balakrishna, Kevin Black, Danny Driess, Michael Equi, Yunhao Fang, Chelsea Finn, Catherine Glossop, Haroun Habeeb, Karol Hausman, Gashon Hussein, Victor Hwang, Brian Ichter, Liyiming Ke, Sergey Levine, Xinyu Li, Yao Lu, Suraj Nair, Karl Pertsch, Allen Z. Ren, Baifeng Shi, Lucy Xiaoyang Shi, Laura Smith, Jost Tobias Springenberg, Kyle Stachowicz, Jiaming Tang, Marcel Torne, Kyle Vedder, Quan Vuong, XuDong Wang, Charles Xu, Lili Yu, Wuming Zhang, Zhuoyang Zhang.

\noindent\textbf{Policy infrastructure.}. Kevin Black, Karan Dhabalia, Danny Driess, Mairbek Khadikov, Chandra Kuchi, Adrian Li-Bell, Vladislav Lialin, Wallace Lim, Yao Lu, Allen Z. Ren, Lucy Xiaoyang Shi, Kyle Stachowicz, Jiaming Tang, Quan Vuong, Haohuan Wang, Ury Zhilinsky.

\noindent\textbf{Robot hardware.}. Ali Amin, Raichelle Aniceto, Greg Balke, Vedant Choudhary, Foster Collins, Grace Connors, Maitrayee Dhaka, Adnan Esmail, Thomas Godden, Ivan Goryachev, Tim Jones, Gregg Kammerer, Ben Katz, Devin LeBlanc, Brendon LeCount, Zhonglin Liang, Enyu Luo, Liam Murphy, Gavin Schelske, Shalom Tekeste, Chris Whalen, Sukwon Yoo.

\noindent\textbf{Robot infrastructure}. Greg Balke, Kevin Black, Shihao Cao, Ken Conley, James Darpinian, Jared DiCarlo, Hunter Hancock, Karol Hausman, Szymon Jakubczak, Jimmy Tanner.

\noindent\textbf{Writing and illustration.}. Bo Ai, Ashwin Balakrishna, Kevin Black, Chelsea Finn, Sergey Levine, Allen Z. Ren, Lucy Xiaoyang Shi, Laura Smith, Kyle Stachowicz.

\subsection{Attention pattern}
\begin{figure}
    \centering
    \includegraphics[width=\linewidth]{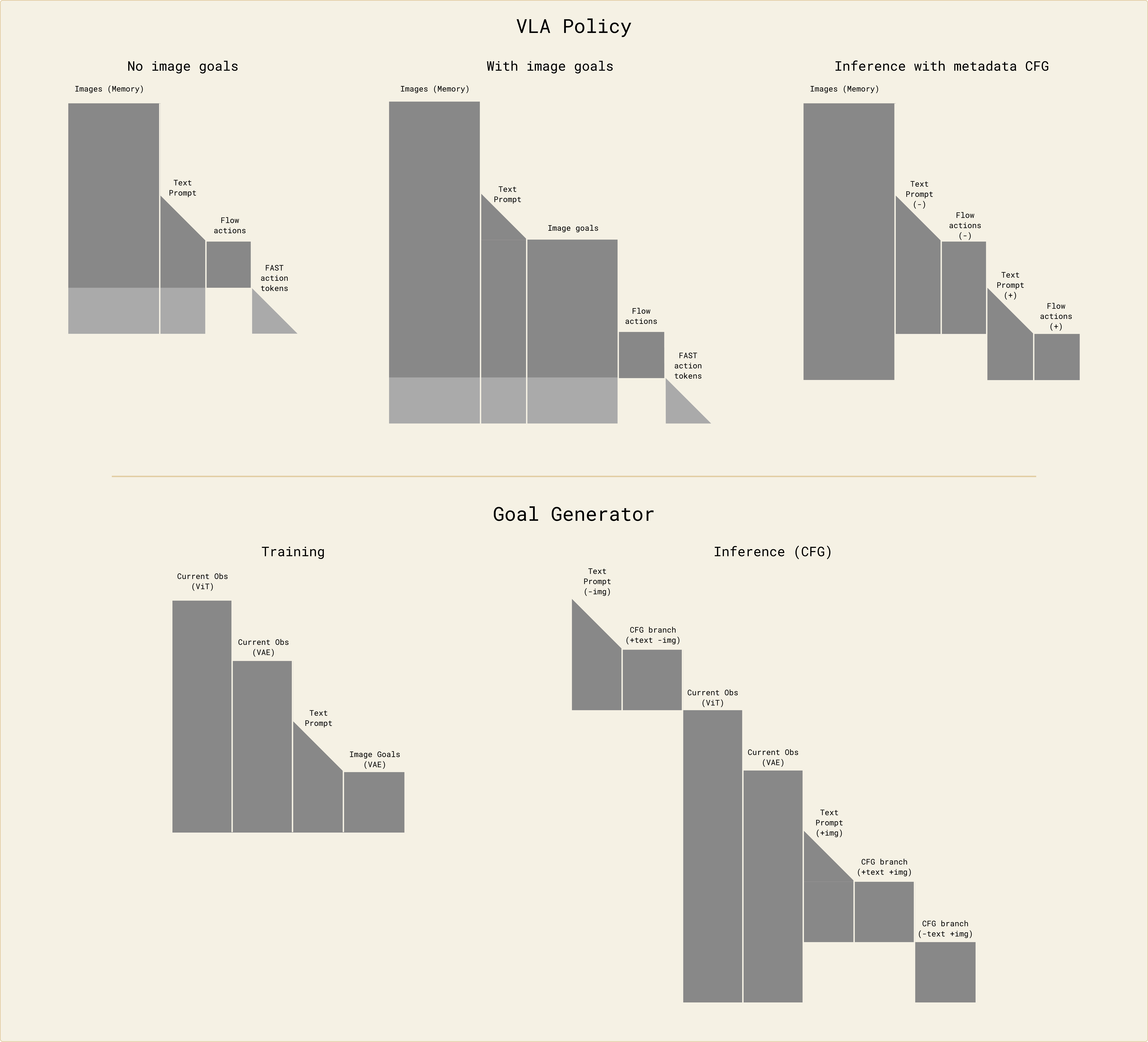}
    \caption{The \MethodName{} model and its world model (for generating subgoal images) use several different nontrivial attention patterns during training and inference. From top left: in absence of image goals we use the same attention patterns as in $\pi_{0.5}$, with global bidirectional attention between embeddings for all memory-aware image views. Note that the FAST tokens (only available at training time) and the flow actions do not attend to each other. When image goals are present, we include them as an additional block-causal bidirectional block, after the text prompt. When we do classifier-free guidance (at inference time), we pack both the positive and negative example into the same sequence for efficient inference by constructing an ``attention tree'' with two branches (positive and negative), which do not attend to one another. Following BAGEL, at training time the world model receives three copies of the image, each of which is block-bidirectional within the multiview group: current observation, encoded with ViT; current observation encoded with VAE; noisy image goal encoded with VAE. The world model uses a similar inference-time CFG trick, albeit with a more complex mask because it has three CFG groups rather than two.}
    \label{app:attention_masks}
\end{figure}

We describe the attention pattern used in training \MethodName{} and the lightweight world model (for subgoal image generation), as well as running inference with them, in Fig.~\ref{app:attention_masks}.

\subsection{Training of the world model}
\label{app:gg_impl}
Our world model is initialized from BAGEL~\citep{deng2025emerging}, and largely uses the same training recipe. We use a subset of our robot data and egocentric human video data with high-quality segmented language labels, as we found that label quality (especially temporal segmentation quality) has a large impact on subgoal quality. We additionally mix in several open-source image editing datasets and open-source video datasets to better preserve the semantic knowledge of the model. Each training example consists of a subtask instruction, $\rawtext$, 3 camera inputs, $\bo_t$, and 3 target images, $\bo_{t_\text{end}}$, where $t_\text{end}$ is the last timestep of the segment spanning $t$. Following BAGEL, the camera inputs are processed using both a ViT (for semantic understanding) and VAE (for fine-grained image details). The ViT tokens are further processed by a 7B LLM backbone, and the VAE tokens are processed by a 7B generation backbone. ViT inputs are resized to a resolution of 448x336, and VAE inputs (including target images) are resized to a resolution of 512x384. The discrepancy is due to the differing patch size of the ViT and VAE (14 and 16, respectively). At test time, we set $\Delta = 4$ seconds as the time interval for regenerating subgoals to match SuSIE~\citep{black2023zero}.

\begin{figure*}[t]
\centering
\includegraphics[width=0.8\linewidth]{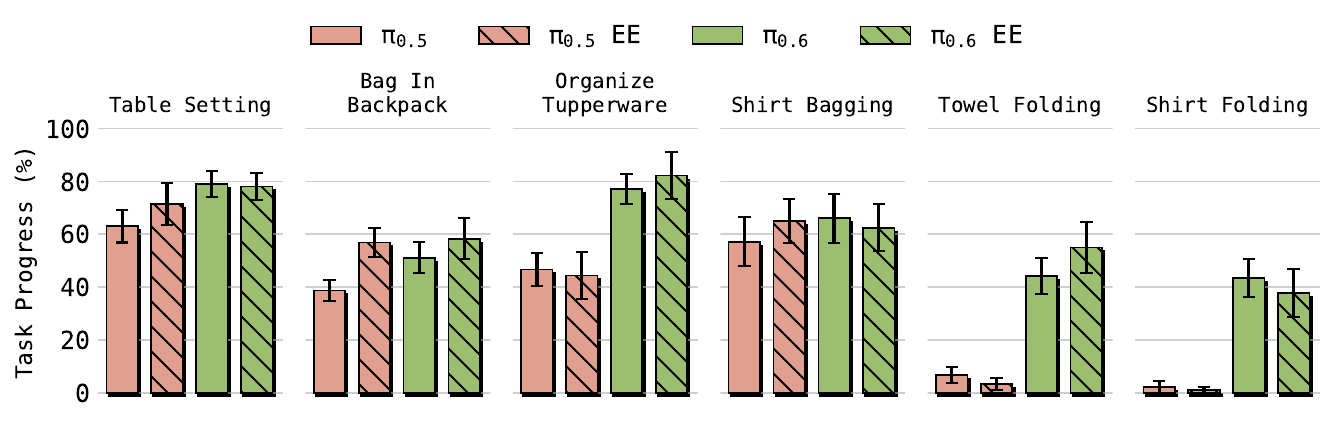}\vspace{-5pt}
\caption{\textbf{Joint vs.\ end-effector control for prior models on cross-embodiment tasks.} We compare joint-space and end-effector (EE) control for baseline policies across a range of tasks, observing no substantial difference in performance between the two control modes.}
\label{fig:xemb_joint_vs_ee}
\end{figure*}

\begin{figure}[t]
    \centering
    \includegraphics[width=0.7\linewidth]{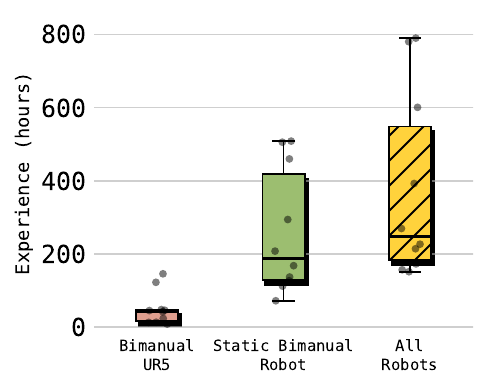}\vspace{-5pt}
    \caption{\textbf{Operator experience in the human subject study.} Box plots show teleoperation experience (in hours) of the ten recruited operators across three categories: UR5e (target robot), the static bimanual robot (source robot), and all robots combined. The selected operators rank within the top 2\% of our operator fleet in terms of teleoperation experience.}
    \label{fig:operator_stats}
\end{figure}

\begin{figure}[t]
    \centering
    \includegraphics[width=0.7\linewidth]{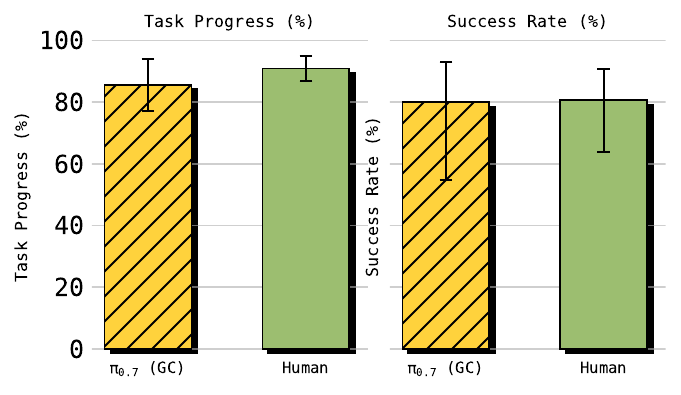}\vspace{-5pt}
    \caption{\textbf{Comparison of \MethodName{} (GC) and human.} We find that \MethodName{} (GC) achieves competitive performance compared to the human operators, in the shirt folding task with the UR5e bimanual platform.}
    \label{fig:human_vs_policy_quantitative}
\end{figure}

\subsection{Inference speed and optimization}
\label{app:inference}
The \MethodName{} model and the high-level policy, which are both based on Gemma3 4B, use a single NVIDIA H100 GPU for inference. Various optimizations implemented after RTC~\citep{black2025real} have brought the inference time of the minimal variant of \MethodName{} down to 38ms with 3 camera inputs, 5 denoising steps, and training-time RTC~\citep{black2025ttrtc} (which, unlike test-time RTC, incurs no additional inference-time overhead). Enabling the \texttt{MEM} vision encoder~\citep{torne2026mem} and adding subgoal images to the context both incur additional overhead, bringing the inference time to 127ms in the worst case.

Generating subgoal images at a reasonable latency is challenging, due to the computational expense of iterative denoising with a 14B model and total sequence length of nearly 10,000 tokens. In addition to the aforementioned optimizations, we also use 4-way tensor parallelism on 4xH100 GPUs, quantize all large matrix multiplications to 8-bit precision, and use a modified version of SageAttention~\citep{zhang2024sageattention} for the backbone attention operations. This allows us to generate subgoal images with 25 denoising steps (each including both text and image CFG) in 1.25 seconds. At inference time, we execute using a naive asynchronous strategy, meaning that \MethodName{} continues executing while the world model generates the next subgoal.

\subsection{Comparison of action spaces in cross-embodiment transfer} \label{app:xemb_joint_vs_ee}

Figure~\ref{fig:xemb_joint_vs_ee} compares joint-space and end-effector (EE) control on cross-embodiment tasks. Across tasks, EE control does not show clear advantage. Therefore, we focus on joint-space control in the main cross-embodiment experiments (Sec.~\ref{sec:cross_emb_transfer}) for clarity. 

\subsection{Human subject study for cross-embodiment shirt folding}  \label{app:human_study}

The human subject study serves two purposes. First, it provides the strongest possible baseline for evaluating \MethodName{} by measuring how well expert operators can teleoperate the UR5e bimanual robot to perform the same task. Second, the result potentially motivates for cross-embodiment transfer: UR5e, an industrial manipulator with high joint inertia, is difficult to teleoperate precisely for dexterous tasks like shirt folding, making demonstration collection on this platform difficult. A model that transfers to new embodiment opens up opportunities to collect autonomous data without human teleoperation.

\textbf{Participant selection.} We recruited ten operators with the top 2 percentile of experience from the entire group of operators. They all have extensive prior experience teleoperating the source static bimanual robot, with an average experience of $\sim$375 hours across all robot platforms (Fig.~\ref{fig:operator_stats}). Crucially, none had prior experience performing the shirt folding task on the UR5e, mirroring the ``zero-shot'' setting of our learned policy.

\textbf{Protocol.} Each operator performed three trials, yielding 30 total trials. To match the zero-shot transfer setting of \MethodName{}, operators received no practice or warm-up period before their first attempt. The initial shirt configuration (flattened on the table), the time limit, and the evaluation criteria were all identical to those used in the \MethodName{} policy evaluation. Operators were instructed to maximize task success to align with the evaluation metric. We report task progress and success rate using the same metrics as in our robot experiments, ensuring a direct and fair comparison.

\textbf{Results.}
Fig.~\ref{fig:human_vs_policy_quantitative} compares \MethodName{} (GC) with human teleoperators under the same zero-shot setting. Human operators achieve an average task progress of 90.9\% and a success rate of 80.6\%. \MethodName{} achieves 85.6\% task progress and an 80\% success rate, demonstrating performance comparable to these expert operators despite not trained with any folding data from the UR5e platforms. These results provide strong evidence of zero-shot cross-embodiment transfer in \MethodName{}.

\subsection{Detailed task descriptions and scoring rubric}
\label{app:tasks}

\noindent \textbf{Laundry (T-shirts and Shorts).} Fold a single pair of shorts or a T-shirt taken from a laundry basket into a compact, neatly aligned fold and place/stack it appropriately.
Scoring: success if successfully folding the item and properly stacking it on top of other laundry.

\noindent \textbf{Laundry (Diverse --- Hardest Item).} Fold a single buttonup shirt taken from a laundry basket into a compact, neatly aligned fold and place/stack it appropriately. Scoring: success if successfully folding the item and properly stacking it on top of other laundry.

\noindent \textbf{Make Espresso.} Execute an espresso workflow to prepare and extract a doppio (double espresso) with correct sequencing and placements.
Scoring: success if completing the required steps in order: grounds dispensed and tamped, portafilter locked into the correct grouphead, extraction performed, cup placed on saucer and moved right.

\noindent \textbf{Box Building.} Assemble a flat box into a 3D box. Scoring: success if the box is folded correctly without major damage.

\noindent \textbf{Make Peanut Butter sandwich.} Make a peanut-butter sandwich, cut it diagonally all the way through, and present it on the plate; close the peanut butter jar; push the plate away.
Maximum score: 9. Points awarded for: removing the jar lid; progressively spreading peanut butter coverage; placing the plain slice on top; cutting diagonally fully through; placing the knife back on the plate; pushing the plate away; and overall neatness.

\noindent \textbf{Turn a T-shirt Inside Out.} Retrieve an inside-out t-shirt, turn it fully right-side-out (torso + both sleeves), fold it into a compact form.
Maximum score: 7. +1 retrieve shirt; +1 torso right-side-out; +1 left sleeve right-side-out; +1 right sleeve right-side-out; +1 compact fold; +1 placed on pile; +1 positioned in upper-right corner.

\noindent \textbf{Drive Through Door.} Open a self-closing closet door, drive the robot fully inside, and finish with the door closed and the robot inside.
Maximum score: 3. +1 for opening the door; +1 for driving far enough inside for the door to close; +1 for smooth entry without bumping into the doorway.

\noindent \textbf{Cut Zucchini.} Slice a zucchini into thin slices using a knife attached to a lanyard while stabilizing the zucchini with the opposite gripper.
Maximum score: 3. +1 for successfully picking up the knife correctly; +1 for fully cutting the zucchini into thin, even slices; +1 for safely returning the knife to the right side of the cutting board.

\noindent \textbf{Peel Fruits and Vegetables.} Peel a fruit/vegetable completely by holding the item against cutting board, and peeling it with the other gripper.
Maximum score: 9. +1 for picking up the peeler; +1 for peeling up to 25\%; +1 for peeling 25–50\%; +1 for peeling 50–75\%; +1 for peeling $>$75\%; +1 for putting the fruit/veg into the bowl; +1 for scraping 25–50\% of food scraps into the trash can; +1 for scraping 50–75\% of food scraps into the trash can; +1 for scraping $>$75\% of food scraps into the trash can.

\noindent \textbf{Take Out Trash.} Remove an old trash bag, place it away from the bin, line the trash can with a new bag, and return the bin to its home location (and close the cabinet / replace lid as appropriate). Maximum score: 12. Taking out trash can (3 points): +2 if the robot correctly accesses the trash can (opens the under-sink cabinet door); +1 if the robot relocates the trash can from under the sink to an accessible position on the kitchen floor. Remove from bin (3 points): +1 if the robot pulls the trash bag away from the corners of the bin; +1 if the robot gathers and securely lifts the trash bag out of the bin; +1 if the robot removes the trash bag from the bin and places it on the floor near the trash can. Replace bag (3 points): +1 if the robot picks up the replacement trash bag; +1 if the robot opens the replacement bag; +1 if the robot places the replacement trash bag fully inside the trash can and stretches the edges securely around the rim. Return trash can back (3 points): +2 if the robot lifts and returns the trash can to the under-sink cabinet; +1 if the robot closes the cabinet door fully at the end of the episode.

\noindent \textbf{Swap 3 Mugs.} Sequentially place three mugs onto a coffee machine drip tray (one at a time), ensuring each mug occupies the tray once, with mugs returned to the table between placements as required. Maximum score: 4. +1 for removing mug 1 from the coffee maker and placing it on the table; +1 for placing mug 2 from the table onto the coffee maker; +1 for removing that same mug 2 from the coffee maker and placing it on the table; +1 for placing the remaining mug 3 from the table onto the coffee maker.

\noindent \textbf{Find Object.} Retrieve an object hidden in a drawer and restore a clean final scene (object on table, drawers closed).
Maximum score: 4. +1 for fully opening the target drawer with the item on the first try; +1 for picking up the sponge stick; +1 for placing the sponge stick on the table (anywhere); +1 for closing the opened target drawer.

\noindent \textbf{Scoop Beans.} Scoop exactly two scoops of coffee beans into a grinder using a measuring cup, then close the grinder lid. Maximum score: 5. +1 for opening the coffee grinder; +1 for grasping the scoop; +1 for successfully collecting and pouring the first full scoop of beans into the coffee grinder; +1 for successfully collecting and pouring the second full scoop of beans into the coffee grinder; +1 for closing the coffee grinder lid.

\noindent \textbf{Window Cleaning.} Spray the phone booth window with windex, rip a paper towel, wipe the glass fully dry, and discard the towel. Maximum score: 5. +1 for spraying the booth; +1 for getting the paper towel for cleaning; +1 for cleaning overall the whole door and leaving it dry; +1 for throwing the paper towel in the trash can; +1 for not leaving any drops left on the door.

\noindent \textbf{Reverse Bussing.} Sort 12 objects with reversed mapping: place trash into the bussing bin and dishes/utensils into the trash can. Maximum score: 12. +1 for successfully putting a plate, cup, bowl, or utensil in the trash (7 total); +1 for successfully putting a plastic bottle, foil, plastic lid, take out container, or chip bag in the bussing bin (5 total).

\noindent \textbf{Reverse Fridge to Microwave.} Move a plate of (real) frozen microwavable food from the microwave into the refrigerator (reverse sequence version), and complete the episode with the plate stored in the fridge.
Maximum score: 6. +1 for microwave door opening; +1 for microwave door closing; +1 for plate removed from microwave; +1 for plate placed in refrigerator; +1 for refrigerator door opening; +1 for refrigerator door closing.

\noindent \textbf{Table Setting.} Set a table by placing a placemat, cup, plates, napkin, and utensils from a bin into a reasonable table setting.
Maximum score: 7. +1 per item successfully placed; critically incorrect placements can be scored as -1.

\noindent \textbf{Bag in Backpack.} Place a small pouch/bag into a backpack (no need to zip closed).
Maximum score: 3. +1 pick up pouch; +1 grasp backpack; +1 place pouch inside backpack.

\noindent \textbf{Organize Tupperware.} Nest three tupperware containers (descending sizes) and stack their corresponding lids.
Maximum score: 6. +1 for each container correctly nested; +1 for each lid correctly stacked/aligned.

\noindent \textbf{Shirt Bagging.} Place two shirts fully inside a brown grocery bag (per-scene), optionally using language commands with timing-sensitive instruction handoff.
Maximum score: 4. +1 for correct pick for each shirt; +1 for each shirt fully placed inside the bag.

\noindent \textbf{Shirt Folding.} Fold a single t‑shirt from a flattened start state into a folded end state.
Maximum score: 6. +1 for completing the first fold (grasp cloth with both arms; folded edge aligned with intended fold within 5 inches); +1 for completing the second fold (same criterion); +1 for completing the final fold (completed with right arm; aligned within 5 inches); plus a fold quality score from 0–3 based on the final folded state chart. Full score of 6 is considered success.

\noindent \textbf{Press French Press Plunger.} Press the french press plunger fully down to the bottom.
Maximum score: 1. +1 if the robot manages to press the plunger down to the bottom (through coffee).

\noindent \textbf{Scoop Rice into Rice Cooker.} Pick up a rice scooper from a rice container, scoop rice, and pour it into an open rice cooker.
Maximum score: 3. +1 pick up scooper; +1 scoop rice; +1 pour into cooker.

\noindent \textbf{Loading an Air Fryer.} Open an air fryer, place a sweet potato into the basket, and close the air fryer.
Maximum score: 4. +1 open air fryer; +1 pick up sweet potato; +1 place into air fryer; +1 close air fryer.

\noindent \textbf{Unloading an Air Fryer.} Pull out an air fryer basket and dump 8 fake fries onto a plate. Maximum score: 2. +1 if the robot successfully picks up all food items from the air fryer; +1 if the robot places all food items onto the plate.

\noindent \textbf{Toast a Bagel.} Place a sliced half-bagel into a toaster oven, initiate toasting by turning the knob, then retrieve and serve it on a plate taken from an overhead cabinet.
Maximum score: 7 (open oven; insert bagel; close oven; turn knob; retrieve plate; retrieve toast; place on plate).

\end{document}